\documentclass{article}

\usepackage{iclr2027_conference,times}

\usepackage[utf8]{inputenc}
\usepackage[T1]{fontenc}
\usepackage{hyperref}
\usepackage{url}
\usepackage{booktabs}
\usepackage{amsmath}
\usepackage{amsfonts}
\usepackage{amssymb}
\usepackage{graphicx}
\usepackage{float}
\usepackage{microtype}
\usepackage{xcolor}

\newcommand{\method}{LayerCanvas}
\newcommand{\dataset}{InfoLayers}

\title{\method: Infographic Generation in AI Learning Assistants \\ via Hybrid Layer-Set Diffusion}

\author{Keshav Gupta$^{1}$\thanks{Work done during an internship at LinkedIn Inc.} \quad Jeremy Owen$^{2}$ \quad Christopher Lloyd$^{2}$ \quad Shaunak Mishra$^{2}$ \quad Manav Sharma$^{2}$ \\
$^{1}$University of California, San Diego \quad $^{2}$LinkedIn Inc.}

\iclrfinalcopy

\begin{document}

\maketitle
\lhead{}\renewcommand{\headrulewidth}{0pt}

\begin{figure}[H]
  \centering
  \includegraphics[width=\linewidth]{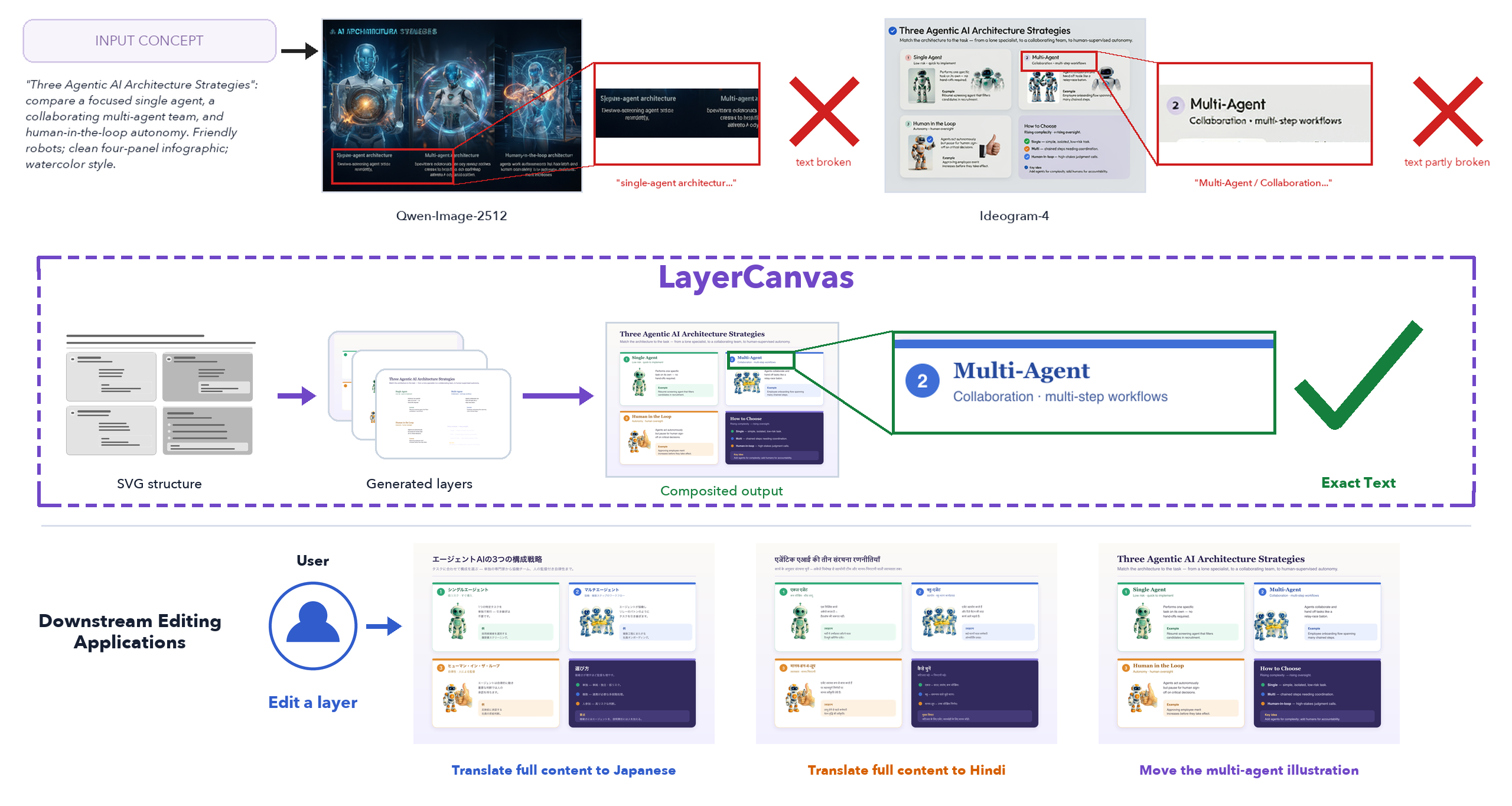}
  \caption{\textbf{\method{} generates editable, text-faithful infographics.}
  Current SOTA text-to-image models such as Qwen-Image-2512~\citep{qwen2025image2512} and
  Ideogram-4~\citep{ideogram2026v4} flatten the graphic and often garble text
  (top). Given a
  vector layout, \method{} generates ordered RGB layers and composes them with exact
  vector text (middle), enabling localization and illustration edits without
  changing unrelated content (bottom).}
  \label{fig:teaser}
\end{figure}

\begin{abstract}
Online learning platforms are increasingly deploying AI assistants to provide personalized, on-demand tutoring. These can leverage infographics to communicate complex information and improve the learning experience. However, automating effective infographic generation remains a challenging task, requiring a combination of visual richness with precise typography, structured layouts, and continued editability.
Existing representation methods satisfy these goals only in part: Scalable Vector Graphics (SVG) representations preserve geometry, hierarchy, and text but provide limited aesthetic flexibility, while raster diffusion models produce richer appearances by flattening content and layout into a single image. Conditioning a diffusion model on a rasterized SVG does not resolve this conflict because the raster preserves approximate placement but discards element ownership, occlusion relationships, and the structure needed to modify and recompose the result. We introduce \emph{LayerCanvas}, a hybrid approach that retains SVG as the semantic, structural, and typographic program while using diffusion to aestheticize non-text content. Rather than generating one flattened raster, LayerCanvas partitions SVG elements into z-ordered sets of mutually non-overlapping elements and processes them through communicating parallel RGB generation streams. The resulting layers are independently editable and deterministically recomposed, while text is rendered from preserved vector metadata. Our approach combines rich visual synthesis with accurate typography and localized editing. LayerCanvas substantially improves text fidelity over state-of-the-art raster image generators while achieving the highest overall aesthetic quality. We further demonstrate non-destructive content editing, localization, asset reuse, and animation. 
\end{abstract}

\section{Introduction}

Visual explanations are a fundamental component of modern learning platforms, where instructional content is increasingly presented through a combination of text, diagrams, illustrations, and infographics. Pairing words with pictures improves comprehension and retention relative to text alone \citep{mayer2009multimedia,clark1991dual}, while infographics can make dense information more accessible and memorable \citep{bateman2010useful,bhat2023infographics}. As learning platforms increasingly deliver and adapt instructional materials digitally, there is a growing need to automatically generate visual explanations that are both pedagogically effective and easy to revise as content evolves. Such explanations must communicate through precise labels and clear visual hierarchy while also providing the visual richness expected from modern image generators \citep{rombach2022ldm,podell2023sdxl,chen2025diffusionsurvey}. Meeting these requirements is challenging: precision, structure, and editability are naturally supported by representations such as Scalable Vector Graphics (SVG)~\citep{w3cSvg11}, whereas visual richness is most readily produced by raster generative models.

% \paragraph{Motivation.}
% This representational contrast has a fundamental tradeoff where vector graphics make geometry, grouping, text, and rendering order explicit, which makes direct manipulation possible \citep{w3cSvg11}. Recent work shows that neural models and increasingly large language models can synthesize such SVG programs directly from text or images \citep{rodriguez2025starvector,wu2023iconshop,xing2024llm4svg,wu2024chat2svg}. However, automatically generated SVGs often appear visually rigid, while handcrafted stylization is costly and difficult to scale. Diffusion and flow models provide richer aesthetic priors \citep{rombach2022ldm,saharia2022imagen,esser2024sd3}, but synthesizing the artifact as a single raster makes exact typography and local intervention difficult; even state-of-the-art image models render dense, legible text unreliably, motivating specialized text-rendering methods \citep{liu2023characteraware,zhang2023textdiffuser,tuo2024anytext}.

This representational contrast has a fundamental tradeoff. Vector graphics make geometry, grouping, text, and rendering order explicit, which makes direct manipulation possible \citep{w3cSvg11}, but automatically generated SVGs \citep{rodriguez2025starvector,wu2023iconshop,xing2024llm4svg,wu2024chat2svg} often appear visually rigid, and stylizing them by hand is costly and difficult to scale. On the other hand, diffusion and flow models provide richer aesthetic priors \citep{rombach2022ldm,saharia2022imagen,esser2024sd3}, but synthesizing the artifact as a single raster makes exact typography and local intervention difficult; even state-of-the-art image models render dense, legible text unreliably, motivating specialized text-rendering methods \citep{liu2023characteraware,zhang2023textdiffuser,tuo2024anytext}. These limitations are especially consequential for learning platforms, which must serve learners worldwide by localizing explanations into different languages and adapting layouts across display formats, from desktop interfaces to mobile screens. Supporting these uses requires text to remain disentangled from imagery and layout components to remain independently adjustable, properties that are lost when the infographic is flattened into a raster. A natural way then is to rasterize the SVG layout and use it as a conditioning signal for a diffusion model, letting the vector structure guide where content goes while diffusion supplies appearance.

Simply rasterizing an SVG for conditioning does not resolve the mismatch: it preserves coarse spatial placement but discards element ownership, occlusion relationships, and the structure needed to recompose the graphic after an edit. This loss of structure cannot be addressed simply by strengthening raster conditioning. A system designed for editable generation should preserve these compositional units throughout synthesis instead of attempting to recover their ownership from pixels afterward.

Motivated by this principle, LayerCanvas combines the structure of SVGs with the visual capabilities of diffusion models. The SVG defines the semantic layout, hierarchy, paint order, and typography. Its non-text elements are partitioned into ordered, non-overlapping layers and generated through communicating parallel RGB streams, which are then alpha-composited according to the SVG z-order. Text nodes are kept outside the diffusion process entirely and rendered directly from their preserved vector metadata (Figure~\ref{fig:teaser}). Outputs remain editable, so users can replace text, move or hide groups, change the z-order, reuse illustrations, or specify animation trajectories. This makes the representation well suited to visual authoring workflows where generated content needs to remain editable and reusable, including educational content creation.

Training such a model requires examples that pair SVG with the corresponding rendered layers and final image. To create this paired supervision, we design a custom multi-agent framework, following recent work on agentic pipelines \citep{jain2026nerfify,huang2026scifig}. The agents create concepts and SVG layouts, add illustrations, produce text-free targets and masks, and check that each example renders and reconstructs correctly.

% \paragraph{Scoped novelty.}
% LayerCanvas's scoped novelty lies in using \emph{renderer-derived, ordered non-overlap layer sets} as the interface between SVG structure and diffusion-based aestheticization. The method assumes that an editable SVG is available and does not replace SVG planning.

Our contributions are:
\begin{itemize}
  \item A hybrid formulation that separates deterministic SVG structure and typography from diffusion-based aestheticization;
  \item A multi-agent pipeline that produces and validates paired SVGs, text-free targets, layers, masks, and provenance metadata;
  \item A rendering-aligned layer representation that supports parallel generation, explicit occlusion modeling, and deterministic z-order composition, yielding over $25\%$ higher placement-exact OCR character accuracy and approximately $27\%$ higher aggregate aesthetic quality than the strongest raster baseline
  % \item An evaluation protocol that measures visual quality, text and structural fidelity, layer reconstruction, edit locality, transformation response, and composability; and
  \item Practical workflows for non-destructive editing, localization, asset reuse, and animated visual explanations.
\end{itemize}

\section{Related Work}
\label{sec:related_work}

\paragraph{Visual documents and typography.}
Automatic graphic design combines templates, constraints, retrieval, and learned layouts; neural methods generate sets or sequences of labeled regions, including alignment and overlap structure \citep{li2019layoutgan,gupta2021layouttransformer,inoue2023layoutdm}.  Such representations organize posters, slides, and documents, but do not by themselves resolve illustration appearance, occlusion, or editable object ownership.  Programmatic graphics preserve structure, yet hand-authored styling and rigid templates scale poorly to open-ended concepts.  Image generators provide richer aesthetics, and text-aware methods introduce character encoders, rendered-glyph conditions, OCR objectives, or localized controls \citep{zhang2023textdiffuser,yang2024glyphcontrol,tuo2024anytext}. Instead, we retain SVG as the source of truth for geometry and typography, remove text from diffusion targets, and render editable vector text after aestheticization.

\paragraph{Controllable, object-centric, and layered generation.}
ControlNet and T2I-Adapter condition pretrained diffusion models on spatial signals \citep{zhang2023controlnet,mou2024t2iadapter}, while grounded, reference-conditioned, and regional methods add box, entity, exemplar, or prompt-region controls \citep{li2023gligen,ye2023ipadapter,bar2023multidiffusion}. These approaches improve spatial control but typically produce a single entangled raster or latent field, limiting direct editing after sampling. Object-centric, compositional, and layer-aware models separate entities, concepts, transparency, or foreground/background content \citep{burgess2019monet,locatello2020objectcentric,avrahami2023breakascene,zhang2024layerdiffusion}, with recent methods explicitly generating or recovering editable image layers \citep{yang2024layerdecomp,yin2025qwenimagelayered,wang2026reveallayer}. In contrast, LayerCanvas uses SVG layer identity as explicit conditioning, packing mutually non-overlapping elements into ordered RGB layer sets. This known layer structure makes output decomposition and local intervention intrinsic to the representation, rather than requiring layers to be recovered post hoc.

% \paragraph{Parallel generation beyond next-token prediction.}
% Diffusion and flow models jointly gradually denoise high-dimensional image states.  Flow matching learns continuous probability paths \citep{lipman2023flowmatching}, rectified flow studies straighter transport trajectories \citep{liu2023flowstraightfast}, and discrete denoising or masked-token models support iterative parallel refinement \citep{austin2021structureddenoising,chang2022maskgit,sahoo2024simpleeffective}.  Our parallel variables are renderer-defined layer streams. Correlated noise promotes global coherence, sparse cross-stream communication coordinates style and occlusion, and z-order composition produces the final artifact.

\paragraph{Agentic multimodal curation and differentiation.}

Foundation models have increasingly been used to generate instructions, propose examples, and curate synthetic multimodal supervision \citep{wang2023selfinstruct,liu2023visualinstruction,gadre2023datacomp}. More recent agentic pipelines extend this paradigm by coordinating specialized agents for scientific tasks, including translating NeRF papers into executable code and automating the generation of editable scientific figures \citep{jain2026nerfify,huang2026scifig}. These systems demonstrate the potential of agentic workflows for generating and iteratively refining structured scientific artifacts. Our pipeline builds on this agentic paradigm by combining specialized generative and validation agents with deterministic SVG parsing, rendering, OCR, overlap analysis, and alpha-composition checks, producing mechanically validated training records.

\section{Problem Setup and Method}
\label{sec:method}

\subsection{Overview}
\label{sec:overview}

Infographics combine two capabilities that are usually separated. SVGs
provide exact layout and typography, whereas
diffusion models provide rich raster appearance but collapse the scene into
one pixel grid \citep{podell2023sdxl,rombach2022ldm}. Conditioning
diffusion on a rasterized SVG preserves coarse
geometry but loses element boundaries and information about which elements
cover others. \method{} instead keeps the SVG as the main scene
description and generates its non-text content as an
ordered set of editable raster layers.

Figure~\ref{fig:layer-diffusion-loop} summarizes the approach. A renderer creates
training layers from the SVG, elements that do not overlap are packed into
ordered groups, and all groups are denoised in parallel by a frozen image
backbone. A trainable adapter identifies each layer and exchanges
messages based on SVG overlap and parent-child links. The generated layers
are then combined in SVG paint order and exact vector text is added back.

\subsection{Problem setup and hybrid output}
\label{sec:problem}

We represent an infographic as an SVG scene
\begin{equation}
    S=(\mathcal{E},\mathcal{T},\mathcal{H},\mathbf{z},\mathcal{M}),
\end{equation}
where \(\mathcal{E}=\{e_i\}_{i=1}^{N}\) contains non-text elements,
\(\mathcal{T}\) stores text and typography, \(\mathcal{H}\) stores
parent-child links, \(z_i\) is the paint index, and \(\mathcal{M}\)
contains element type \(\tau_i\), normalized bounding box \(b_i\), and
renderer-derived masks. In particular, \(A_i\) is the full (amodal) mask of
element \(e_i\), while \(V_i\) is the part that remains visible after
later elements cover it. Keeping both masks provides training targets for
content that becomes visible after deletion, translation, or z-order edits.

The output is a hybrid artifact,
\begin{equation}
    \mathcal{A}=\left(S,\{\widehat L_g\}_{g=1}^{G},R\right),\qquad
    \widehat I
    =\operatorname{Over}\!\left(
      R_{\rm text}(\mathcal{T}),
      \operatorname{Comp}_{z}(\widehat L_1,\ldots,\widehat L_G)\right),
    \label{eq:artifact}
\end{equation}
where \(\widehat L_g=(\widehat C_g,\widehat\alpha_g)\) is an editable RGB
layer and \(\operatorname{Comp}_z\) composes layers in paint order. Diffusion
therefore generates the non-text appearance, while the SVG continues to
define geometry, ownership, which elements cover others, and exact text.

% An offline renderer supplies text-free targets, per-element RGB assets,
% full and visible masks, types, boxes, parent-child links, and paint order.
% Thus, layer ownership is computed directly from the SVG rather than inferred
% from image pixels. Further details about the training targets and
% text are given in Appendix~\ref{app:method-details}.

\subsection{Ordered non-overlapping layer sets}
\label{sec:grouping}

Assigning one diffusion path to every SVG element preserves ownership but
is unnecessarily expensive: many elements never overlap and can share a
path without competing for pixels. Unlike pixel-derived layer decomposition
\citep{zhang2024layerdiffusion}, we use the SVG's explicit element ownership
to form ordered groups
\(\mathcal{G}=(G_1,\ldots,G_G)\) of \(\mathcal{E}\). Elements within a group
have non-overlapping full masks,
\begin{equation}
    A_i\odot A_j=0,\qquad i\ne j,\quad e_i,e_j\in G_g,
    \label{eq:nonoverlap}
\end{equation}
while every overlap respects paint order: if \(A_i\odot A_j\ne0\) and
\(z_i<z_j\), then \(g(i)<g(j)\). Consequently, non-overlapping elements may
share one generated layer, but an element in front and the element behind it remain
separately editable.

We create the groups by processing elements in ascending \((z_i,i)\)
order:
\begin{equation}
 g(i)=
 \begin{cases}
  1, & \mathcal{P}_i=\varnothing,\\
  1+\max_{j\in\mathcal{P}_i}g(j), & \text{otherwise},
 \end{cases}
 \quad
 \mathcal{P}_i=\{j:z_j\le z_i,\ A_j\odot A_i\ne0,\ j\text{ processed}\}.
 \label{eq:ordered-coloring}
\end{equation}
This rule cannot place a front element behind an element that it covers.
Group masks are
unions, \(A_g=\max_{i\in G_g}A_i\) and
\(V_g=\max_{i\in G_g}V_i\), and each layer path receives its grouped control
raster, type, box, paint order, and owned illustration prompt. The number of paths
therefore depends on how many layers overlap, rather than the number of SVG
elements.

For premultiplied color \(P_g=\alpha_g C_g\), the final raster is obtained by
back-to-front alpha composition,
\begin{equation}
 (P^{(g)},\alpha^{(g)})=
 \big(P_g+(1-\alpha_g)P^{(g-1)},
      \alpha_g+(1-\alpha_g)\alpha^{(g-1)}\big),
 \label{eq:alpha}
\end{equation}
starting from a transparent canvas. Paint order controls composition, not
generation order: all valid groups are denoised in parallel, and individual
layers can later be translated, removed, or reordered without regenerating
unrelated groups.

\subsection{Generating coordinated layers}
\label{sec:architecture}

\method{} generates all layers together, but separate layers can easily drift
toward different colors, lighting, or object identities. Following the
continuous interpolation used in flow matching \citep{lipman2023flowmatching},
we give
them part of the same starting noise and use the same denoising timestep.
This encourages a common look while preserving private noise for
layer-specific detail. Shared noise cannot make one layer respond to another,
however, so the frozen image backbone needs a small trainable adapter. The
adapter follows the efficient frozen-backbone conditioning strategy of
ControlNet and T2I-Adapter \citep{zhang2023controlnet,mou2024t2iadapter}, but
operates across an ordered layer set so that each layer is influenced by the other. It answers three questions: which layer is being generated, which other
layers matter to it, and whether their effect should be global or limited to
a shared boundary.

\begin{figure*}[t]
  \centering
  \includegraphics[width=\linewidth]{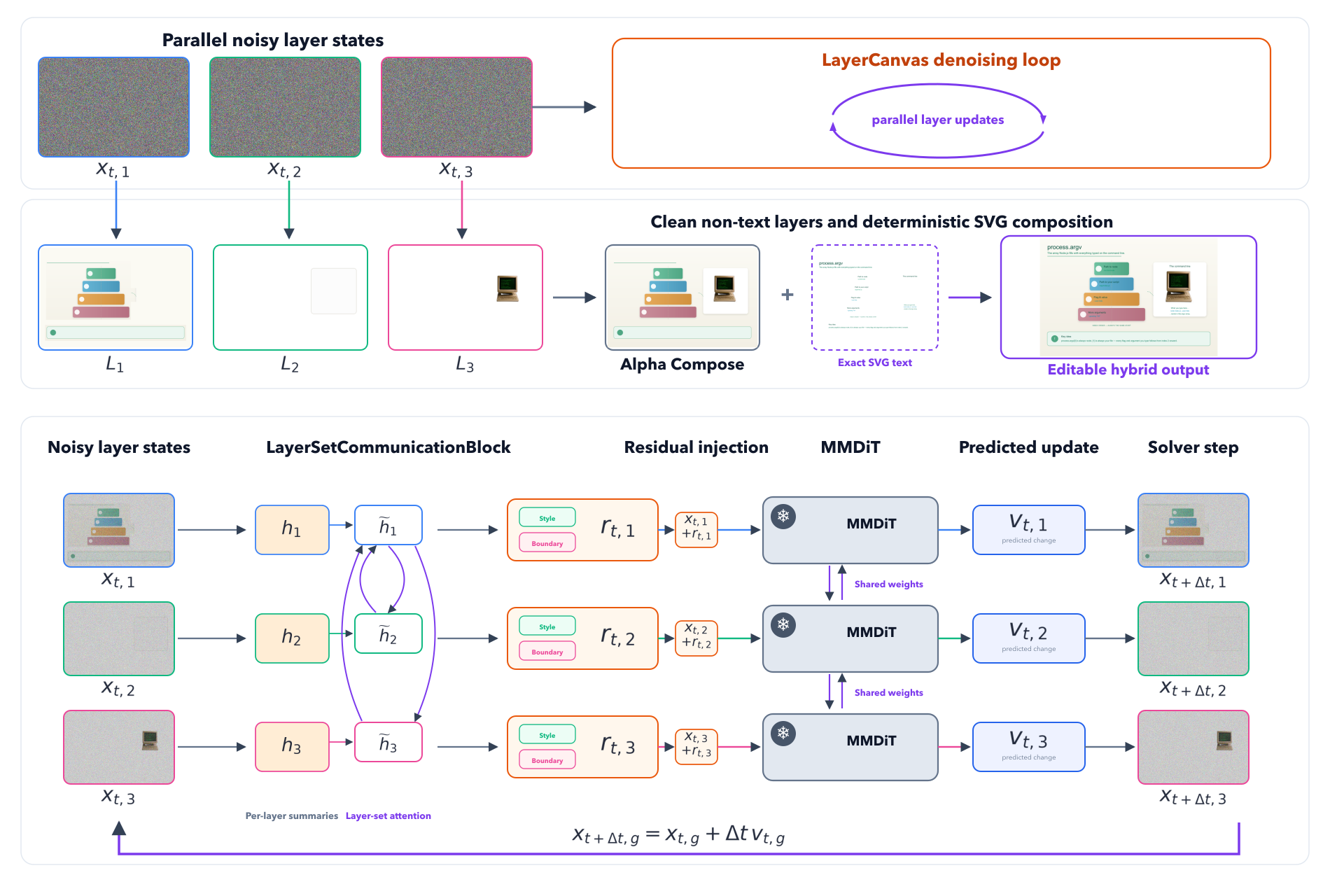}
  \caption{\textbf{\method{} layer generation.} \textbf{Top:} Parallel
  denoising produces aligned non-text layers, which are composed in paint
  order before exact SVG text is restored. \textbf{Bottom:} Layer-set
  attention produces residuals for shared frozen MMDiT updates, and the
  solver advances every layer.}
  \label{fig:layer-diffusion-loop}
\end{figure*}

\paragraph{Identifying each layer.}
The frozen backbone normally generates one image and has no reason to know
whether a batch item represents a background, a foreground illustration, or
a later paint-order layer. For group \(g\), the model's variational
autoencoder (VAE)~\citep{kingma2013auto}
converts the grouped control raster into internal features \(s_g\). The
trainable grouped layer-control adapter then adds learned identifiers for paint
order and the main element type:
\begin{equation}
 q_g=\operatorname{LN}(s_g)+E_z[g]+E_\tau[\tau_g],\qquad
 r_g^{(0)}=W_0q_g,\quad r_g^{(d)}=W_d\phi(W_sq_g).
 \label{eq:layer-control}
\end{equation}
Here, \(\operatorname{LN}(s_g)\) describes the layer's visual control,
\(E_z[g]\) tells the model where it appears in paint order, and
\(E_\tau[\tau_g]\) tells it the main element type. The projections
\(W_0\) and \(W_d\) turn this combined description into small adjustments at
the transformer input and selected depths. These outputs start at zero, so
the adapter initially leaves the pretrained backbone unchanged. The
adjustments are added at a small set of backbone depths.

\paragraph{Sharing information between layers.}
Layer identifiers solve ownership, but the backbone still processes every
layer independently. Dense communication between every image location would
be costly and would ignore the overlap and parent-child links already
available in the SVG. Instead, each
\emph{LayerSetCommunicationBlock} builds one compact summary \(h_g\) per
layer. For a receiving layer \(g\), a sending layer \(h\), and attention head
\(a\), the block first computes a pairwise score \(\ell_{gh}^{a}\), converts
the scores into weights \(\pi_{gh}^{a}\), and uses those weights to combine
the sending-layer summaries:
\begin{equation}
 \begin{aligned}
 \ell_{gh}^{a}
 &=\frac{\langle W_Q^ah_g,W_K^ah_h\rangle}{\sqrt{d_a}}
   +(w_E^a)^\top E_{gh},\\
 \pi_{gh}^{a}
 &=\operatorname{softmax}_{h}(\ell_{gh}^{a}),\\
 \widetilde h_g
 &=W_O\operatorname{Concat}_{a}\!
   \left(\sum_h \pi_{gh}^{a}W_V^ah_h\right).
 \end{aligned}
 \label{eq:graph-attention}
\end{equation}
The first part of \(\ell_{gh}^{a}\) measures similarity between the two
layers; the second adds their SVG relationship from \(E_{gh}\).
\(\pi_{gh}^{a}\) says how much layer \(g\) should listen to layer \(h\), and
\(\widetilde h_g\) is the resulting information received by layer \(g\)
after all attention heads are combined. Because the SVG links have
directions, a covered background can receive a different message from the
element in front of it.

\paragraph{Applying shared information.}
Not every cross-layer change should affect the same area. A shared palette or
lighting choice should act smoothly across a layer, while an overlap
correction should stay near the touching edges. We therefore split the
received summary \(\widetilde h_g\) into a style path and a boundary path.
Let \(x_g\) denote the current spatial features of layer \(g\):
\begin{equation}
 \begin{aligned}
 r_g^{\rm sty}&=F_{\rm sty}(\widetilde h_g),\\
 r_g^{\rm bnd}&=F_{\rm bnd}(x_g,\widetilde h_g),\\
 r_g&=\sigma(\gamma_g^{\rm sty})\,r_g^{\rm sty}
     +\sigma(\gamma_g^{\rm bnd})\,
       B_g\odot r_g^{\rm bnd}.
 \end{aligned}
 \label{eq:branches}
\end{equation}
The style function turns \(\widetilde h_g\) into a smooth layer-wide change.
The boundary function combines \(\widetilde h_g\) with the current features
\(x_g\), while \(B_g\) keeps its change near layer edges. The learned gates
\(\gamma_g^{\rm sty}\) and \(\gamma_g^{\rm bnd}\) decide how strongly each
path should act. Their weighted sum is the final residual \(r_g\), which is
added back to layer \(g\) at that depth of the frozen MMDiT image backbone. As above, the added outputs
start at zero, so communication is introduced gradually during training.

\paragraph{Training and exact text.}
Training checks each layer, the final combined image, hidden areas and 
boundaries. These
losses check both individual ownership and the assembled image.
Typography is excluded from diffusion targets and added back from the retained
SVG strings.
This guarantees exact editable text rather than asking diffusion to paint
glyphs, as in text-aware raster generators
\citep{zhang2023textdiffuser,tuo2024anytext}.
Appendix~\ref{app:method-details} gives the full training path, loss and text-composition procedure.

\begin{table*}[t]
  \caption{\textbf{Primary comparison} on test set. Char and Exact
  measure placement-exact OCR, while Recall measures text recovered anywhere in
  the image. Appeal, Typog., and Cov. are VLM scores from 0-5, with Avg denoting
  their mean. Best synthesis result in each column is \textbf{bold}.}
  \label{tab:main}
  \centering
  \small
  \setlength{\tabcolsep}{5pt}
  \begin{tabular}{lccccccc}
    \toprule
    & \multicolumn{3}{c}{Text fidelity} & \multicolumn{4}{c}{VLM judge (0-5)} \\
    \cmidrule(lr){2-4}\cmidrule(lr){5-8}
    Method & Char & Exact & Recall & Appeal & Typog. & Cov. & Avg \\
    \midrule
    SVG (LLM-authored) & 0.856 & 0.676 & 0.964 & 2.37 & 4.02 & 4.90 & 3.76 \\
    \midrule
    Qwen-Image-2512~\citep{qwen2025image2512} & 0.080 & 0.002 & 0.149 & 2.69 & 1.00 & 1.45 & 1.71 \\
    Ideogram-4~\citep{ideogram2026v4} & 0.657 & 0.360 & 0.744 & 2.92 & 2.29 & 3.89 & 3.03 \\
    \method{} (Ours) & \textbf{0.822} & \textbf{0.623} & \textbf{0.942}
      & \textbf{2.99} & \textbf{3.81} & \textbf{4.75} & \textbf{3.85} \\
    \bottomrule
  \end{tabular}
\end{table*}

\begin{figure*}[t]
  \centering
  \includegraphics[width=\linewidth]{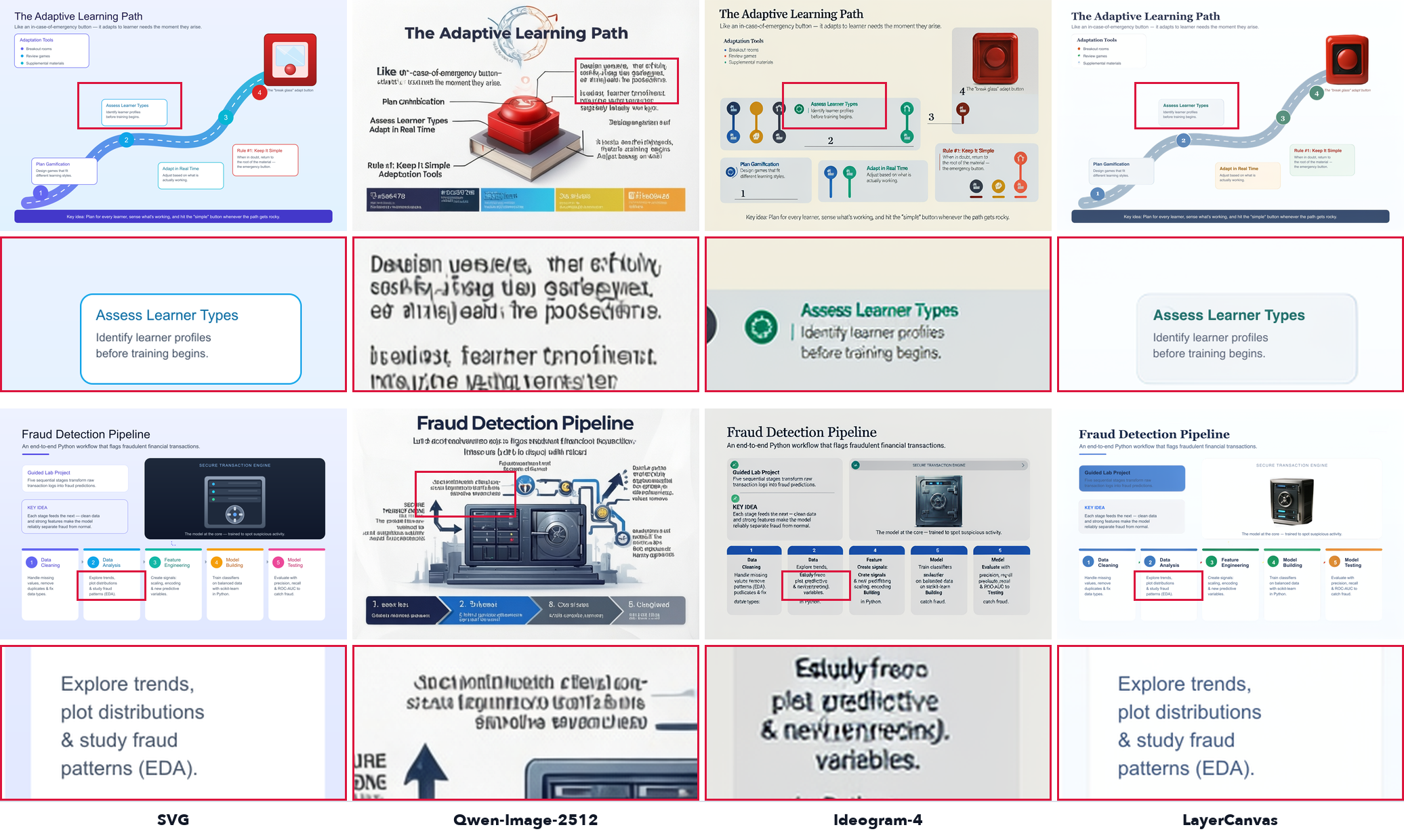}
  \caption{\textbf{Qualitative four-way comparison}
  Columns show, left to right, the SVG control, two prompt-only SOTA
  diffusion baselines (Qwen-Image-2512 \cite{qwen2025image2512} and Ideogram-4 \cite{ideogram2026v4}), and \method{}. The corresponding next row shows the zoomed in regions for the highlighted text regions.
  Prompt-only models produce appealing raster illustrations but garble or
  invent text and drift from the prescribed layout whereas \method{} keeps the layer layout and composites exact, legible vector text.}
  \label{fig:qual_4way}
\end{figure*}

\section{Experiments}
\label{sec:experiments}

% We evaluate whether renderer-derived layer sets preserve visual quality while
% delivering exact text, faithful structure, and localized editing.  Our primary
% comparison places \method{} against production text-to-image systems and a
% deterministic SVG reference under matched prompts, seeds, and sampling, and is
% reported here together with the representation ablation.
% Controlled-transformation study and the
% systems profile are deferred to Appendices~\ref{app:transformations} and~\ref{app:systems-profile}.

\subsection{Setup}
\label{sec:exp_setup}

\paragraph{Dataset -}
We develop \dataset{}, a corpus of 23{,}960 paired infographic records on a
$768\times512$ canvas with a multi-agent pipeline.  Each sample pairs a structural SVG with a text-free
and a composited target, together with the per-layer supervision our model
consumes: amodal and visible masks, paint order, and a hierarchy/overlap graph. Construction and layout statistics are given in
Appendix~\ref{app:data}.

\paragraph{Baselines -}
We compare four methods on shared held-out layouts.  The \emph{LLM-authored
SVG} reference recolors and completes the structural SVG and renders it, giving
exact vector text by construction but no learned aestheticization.  The two
prompt-only open-weight SOTA diffusion baselines, \emph{Qwen-Image-2512}~\citep{qwen2025image2512}
and \emph{Ideogram-4}~\citep{ideogram2026v4},
are each conditioned on the design-concept prompt derived from the input SVG layout; for Ideogram-4 this prompt is
constructed from the same SVG layout, with its structural elements serialized
into the structured prompt format the model expects.  \method{} is our grouped
layer-set diffusion model with deterministic exact-SVG text compositing with the MMDiT backbone as \cite{ideogram2026v4}.
For evaluation, all methods are placed on a $768{\times}512$ canvas.

\paragraph{Metrics -}
Text fidelity is scored two ways.  \emph{Placement-exact} fidelity uses
ground-truth-region OCR \cite{tesseractocr2026} at $768{\times}512$, and reports character accuracy and exact-match rate.
\emph{Placement-agnostic} fidelity runs full-image OCR at the same resolution and
measures word recall, how much of the ground-truth text appears anywhere in the
image, so a baseline that renders readable text at the wrong position still gets
credit.  The two exact-text methods place
text from the SVG by construction, so their remaining gap from $1.0$ is OCR
error rather than real text mistakes. For visual quality we use a \emph{VLM as a judge}:
Claude Opus 4.8 \citep{anthropic2026claudeopus48} scores each design from $0$ to $5$ on illustration appeal,
typography, and semantic coverage, and we report \emph{Avg}, the mean of the
three.  The full evaluation protocol is given in
Appendix~\ref{app:evaluation-details}.

\subsection{Main comparison}
\label{sec:exp_main}

Table~\ref{tab:main} reports the comparison on the test set (construction in
Appendix~\ref{app:additional-experiments}).  \method{} combines strong visual
appeal with text fidelity, outperforming both raster baselines across all text
metrics while approaching the exact-text SVG reference. Ideogram-4 \cite{ideogram2026v4} generates aesthetic looking samples, but it still trails \method{} on text fidelity;
Qwen-Image-2512 \cite{qwen2025image2512} performs poorly on both text placement and recovery. Under the primary
Claude judge, \method{} is the strongest learned method across illustration
appeal, typography, semantic coverage, and aggregate quality, and slightly
exceeds the SVG reference overall.
The same learned-method ranking holds under independent Gemini 3.7 Flash
\citep{deepmind2026gemini37flash} and GPT-5.6 Sol
\citep{openai2026gpt56sol} judges (Appendix
Table~\ref{tab:multi-judge-robustness}). Figure~\ref{fig:qual_4way}
shows the same appeal-text trade-off qualitatively. Additional results are in Appendix~\ref{app:additional-comparisons}. Inference compute details for \method{} are in Appendix~\ref{app:inference-compute}.

\subsection{Representation ablation}
\label{sec:exp_ablation}

\begin{figure*}[t]
  \centering
  \includegraphics[width=\textwidth]{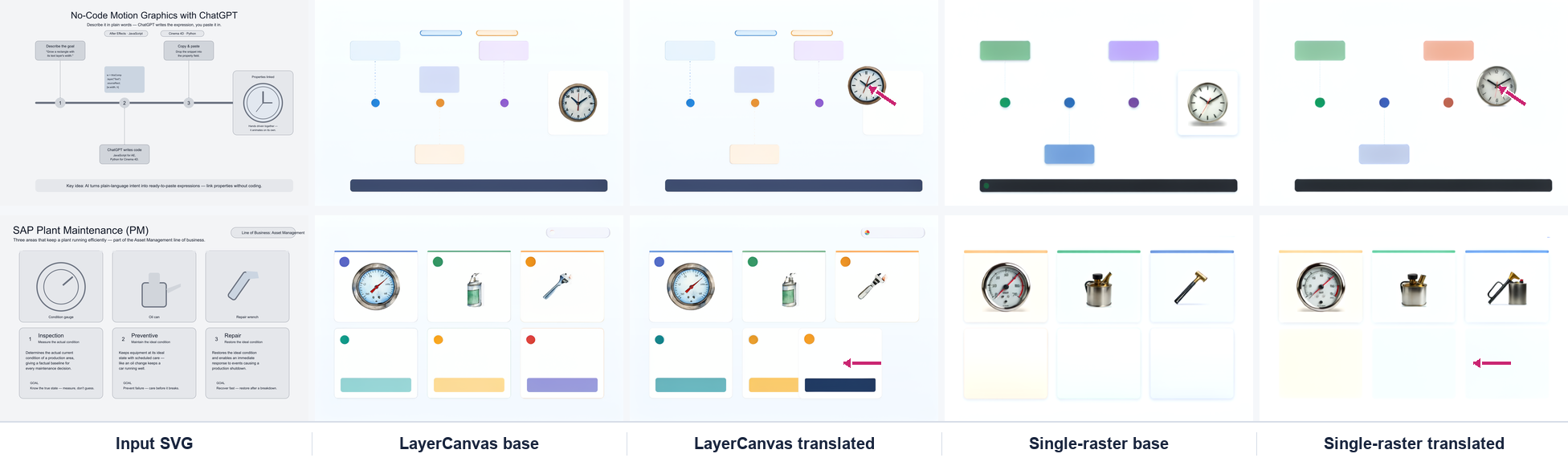}
  \caption{\textbf{Representation ablation:} Each column shows the source
  SVG, fixed-noise base generation, and the translated version by perturbing the base SVG for \method{} and the single-raster as control. Red arrows show the shift from the
  base position in the translated cases. \method{} better preserves object
  identity and unrelated layout content.}
  \label{fig:qualitative_edits}
\end{figure*}

We compare grouped \method{} with a single-raster conditioning baseline to test whether
explicit layer separation improves editability. The baseline conditions the
same MMDiT model \citep{ideogram2026v4} directly on the rasterized SVG using
T2I-Adapter \citep{mou2024t2iadapter}. A single stream must learn to render the
whole graphic at once, which becomes harder as the layout grows denser. In
contrast, each \method{} layer contains non-overlapping elements, making it
simpler to learn before the layers are composed. Under fixed noise, \method{}
better preserves the moved element and surrounding layout, while the control
redraws unrelated content (Figure~\ref{fig:qualitative_edits}). It also improves
text recall and all VLM scores, raising average quality from 3.55 to 3.77
(Table~\ref{tab:single-raster-conditioning}).

\begin{table}[t]
  \centering
  \caption{\textbf{Comparison of Different Conditioning Representations on validation set}}
  \label{tab:single-raster-conditioning}
  \small
  \begin{tabular}{lccccccc}
    \toprule
    Method & Char & Exact & Recall & Appeal & Typog. & Cov. & Avg \\
    \midrule
    Single-raster Conditioning & \textbf{0.851} & \textbf{0.568} & 0.783 & 2.78 & 3.56 & 4.32 & 3.55 \\
    \method{} & 0.823 & 0.530 & \textbf{0.795} & \textbf{2.84} & \textbf{3.76} & \textbf{4.72} & \textbf{3.77} \\
    \bottomrule
  \end{tabular}
\end{table}

% \paragraph{Compositional and edit evaluation.}
% Beyond this static comparison, we also check how well the generated layers
% behave.  We evaluate the decoded RGB streams and their ordered composite by
% group count, overlap, hierarchy, and illustration count, and we test localized
% fixed-noise edits such as translation, valid z-order swaps, deletion, insertion,
% and text replacement, which check whether an edit moves only the intended
% content while clearing its source and revealing the structure behind it. 
\section{Downstream applications}
\label{sec:downstream}

Because \method{} preserves SVG structure alongside independently addressable
raster layers, a generated design can be adapted without regenerating the whole
image. Vector text supports changes to wording, typography, and language, while
separate visual layers enable animation, asset reuse, and local element edits such as removing a card or swapping an
icon while preserving unaffected content.
Retained grouping and paint order further support layout retargeting, re-theming,
and SVG-level corrections while limiting changes to the intended components.
Together, these capabilities turn a generated infographic into a reusable
authoring artifact. We highlight three below; the rest are detailed in
Appendix~\ref{app:downstream-applications}.

\subsection{Text editing and localization}

End-to-end raster generators must repaint text whenever wording, typography, or
language changes. \method{} instead generates only the visual substrate and
preserves typography as styled SVG nodes, so copy, style, and locale changes
require no diffusion pass and cannot alter the generated artwork
(Figure~\ref{fig:downstream-text-editing-localization}).

\subsection{Independent layer animation}

We adapt the zero-shot animation principle of
Text2Video-Zero~\citep{khachatryan2023text2videozero}, in which the motion and
appearance priors of a pretrained text-to-image diffusion model produce a
temporally linked sequence without video-specific training. Whereas a flattened
document would expose the complete page to this update, LayerCanvas lets the
temporal operation target one named image layer while all other layers
remain fixed: every unselected image layer and the full-resolution SVG
typography are copied identically into every frame.

\begin{figure}[t]
  \centering
  \includegraphics[width=\linewidth]{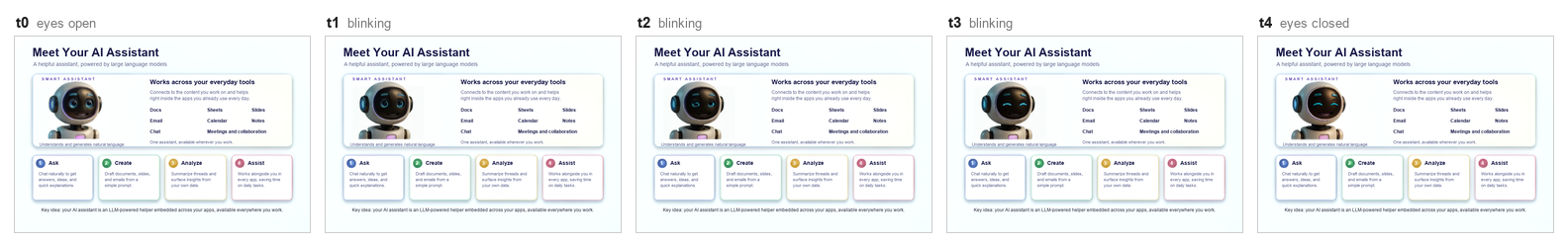}
  \caption{\textbf{Independent animation of one LayerCanvas region.} A
  Text2Video-Zero-inspired image-diffusion trajectory produces an
  open-to-closed blink. Only the selected eye region changes across the five
  frames; the surrounding illustration, layout, and vector typography remain
  fixed.}
  \label{fig:downstream-animation-progressive-reveal}
\vspace{-0.12in}
\end{figure}

\subsection{SVG-space refinement and faithful diffusion propagation}

An upstream layout proposal may contain a misplaced, mislabeled, or incorrectly
styled element. In \method{}, the SVG is a persistent control surface:
the element can be corrected directly, and the model propagates that correction
to the raster without retraining or redesigning the rest of the document.

\begin{figure}[t]
  \centering
  \includegraphics[width=0.88\linewidth]{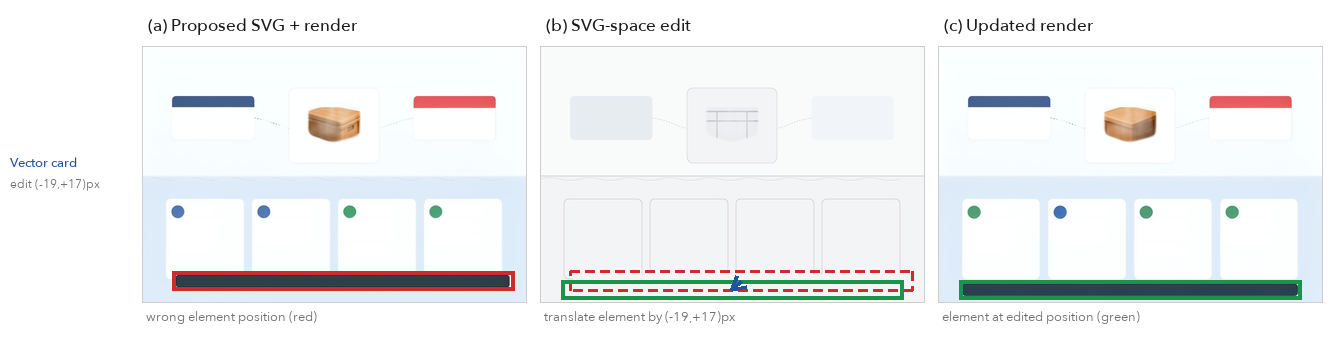}
  \caption{\textbf{Closed-loop SVG correction.} Each row shows an incorrect SVG
  and render, the direct SVG edit, and the corrected render. The edited element
  moves to the requested location while unrelated content remains stable.}
  \label{fig:downstream-svg-refinement-propagation}
\vspace{-0.12in}
\end{figure}

\section{Conclusion}

We formulate infographic generation as structured aestheticization: a deterministic SVG supplies geometry, hierarchy, typography, while parallel layer-set generation supplies visual appearance.  Aligning this representation with the renderer supports many downstream applications like editing and artifact reuse. Enforcing generalization to more layouts and improving transformations under changed occlusion remain important research directions.

\bibliography{references}

@inproceedings{khachatryan2023text2videozero,
  title     = {Text2Video-Zero: Text-to-Image Diffusion Models are Zero-Shot Video Generators},
  author    = {Khachatryan, Levon and Movsisyan, Andranik and Tadevosyan, Vahram and
               Henschel, Roberto and Wang, Zhangyang and Navasardyan, Shant and
               Shi, Humphrey},
  booktitle = {Proceedings of the IEEE/CVF International Conference on Computer Vision (ICCV)},
  pages     = {15954--15964},
  year      = {2023},
  doi       = {10.1109/ICCV51070.2023.01462}
}

@inproceedings{li2019layoutgan,
  title = {LayoutGAN: Generating Graphic Layouts with Wireframe Discriminators},
  author = {Li, Jianan and Yang, Jimei and Hertzmann, Aaron and Zhang, Jianming and Xu, Tingfa},
  booktitle = {International Conference on Learning Representations},
  year = {2019}
}

@inproceedings{gupta2021layouttransformer,
  title = {LayoutTransformer: Layout Generation and Completion with Self-Attention},
  author = {Gupta, Kamal and Lazarow, Justin and Achille, Alessandro and Davis, Larry S. and Mahadevan, Vijay},
  booktitle = {IEEE/CVF International Conference on Computer Vision},
  year = {2021}
}

@inproceedings{inoue2023layoutdm,
  title     = {LayoutDM: Discrete Diffusion Model for Controllable Layout Generation},
  author    = {Inoue, Naoto and Kikuchi, Kotaro and Simo-Serra, Edgar and
               Otani, Mayu and Yamaguchi, Kota},
  booktitle = {Proceedings of the IEEE/CVF Conference on Computer Vision and Pattern Recognition (CVPR)},
  year      = {2023},
  pages     = {10167--10176}
}

@inproceedings{zhang2023textdiffuser,
  title = {TextDiffuser: Diffusion Models as Text Painters},
  author = {Chen, Jingye and Huang, Yupan and Lv, Tengchao and Cui, Lei and Chen, Qifeng and Wei, Furu},
  booktitle = {Advances in Neural Information Processing Systems},
  year = {2023}
}

@inproceedings{yang2024glyphcontrol,
  title     = {GlyphControl: Glyph Conditional Control for Visual Text Generation},
  author    = {Yang, Yukang and Gui, Dongnan and Yuan, Yuhui and Liang, Weicong and
               Ding, Haisong and Hu, Han and Chen, Kai},
  booktitle = {Advances in Neural Information Processing Systems},
  volume    = {36},
  year      = {2024}
}

@inproceedings{tuo2024anytext,
  title     = {AnyText: Multilingual Visual Text Generation and Editing},
  author    = {Tuo, Yuxiang and Xiang, Wangmeng and He, Jun-Yan and Geng, Yifeng and Xie, Xuansong},
  booktitle = {International Conference on Learning Representations},
  year      = {2024}
}

@inproceedings{zhang2023controlnet,
  title = {Adding Conditional Control to Text-to-Image Diffusion Models},
  author = {Zhang, Lvmin and Rao, Anyi and Agrawala, Maneesh},
  booktitle = {IEEE/CVF International Conference on Computer Vision},
  year = {2023}
}

@inproceedings{mou2024t2iadapter,
  title     = {T2I-Adapter: Learning Adapters to Dig Out More Controllable Ability for Text-to-Image Diffusion Models},
  author    = {Mou, Chong and Wang, Xintao and Xie, Liangbin and Wu, Yanze and
               Zhang, Jian and Qi, Zhongang and Shan, Ying},
  booktitle = {Proceedings of the AAAI Conference on Artificial Intelligence},
  volume    = {38},
  number    = {5},
  pages     = {4296--4304},
  year      = {2024}
}

@inproceedings{li2023gligen,
  title = {GLIGEN: Open-Set Grounded Text-to-Image Generation},
  author = {Li, Yuheng and Liu, Haotian and Wu, Qingyang and Mu, Fangzhou and Yang, Jianwei and Gao, Jianfeng and Li, Chunyuan and Lee, Yong Jae},
  booktitle = {IEEE/CVF Conference on Computer Vision and Pattern Recognition},
  year = {2023}
}

@article{ye2023ipadapter,
  title = {IP-Adapter: Text Compatible Image Prompt Adapter for Text-to-Image Diffusion Models},
  author = {Ye, Hu and Zhang, Jun and Liu, Sibo and Han, Xiao and Yang, Wei},
  journal = {arXiv preprint},
  year = {2023}
}

@inproceedings{bar2023multidiffusion,
  title = {MultiDiffusion: Fusing Diffusion Paths for Controlled Image Generation},
  author = {Bar-Tal, Omer and Yariv, Lior and Lipman, Yaron and Dekel, Tali},
  booktitle = {International Conference on Machine Learning},
  year = {2023}
}

@article{burgess2019monet,
  title = {MONet: Unsupervised Scene Decomposition and Representation},
  author = {Burgess, Christopher P. and Matthey, Loic and Watters, Nicholas and Kabra, Rishabh and Higgins, Irina and Botvinick, Matthew and Lerchner, Alexander},
  journal = {arXiv preprint},
  year = {2019}
}

@inproceedings{locatello2020objectcentric,
  title = {Object-Centric Learning with Slot Attention},
  author = {Locatello, Francesco and Weissenborn, Dirk and Unterthiner, Thomas and Mahendran, Aravindh and Heigold, Georg and Uszkoreit, Jakob and Dosovitskiy, Alexey and Kipf, Thomas},
  booktitle = {Advances in Neural Information Processing Systems},
  year = {2020}
}

@inproceedings{avrahami2023breakascene,
  title     = {Break-A-Scene: Extracting Multiple Concepts from a Single Image},
  author    = {Avrahami, Omri and Aberman, Kfir and Fried, Ohad and
               Cohen-Or, Daniel and Lischinski, Dani},
  booktitle = {SIGGRAPH Asia 2023 Conference Papers},
  year      = {2023},
  articleno = {96},
  numpages  = {12},
  doi       = {10.1145/3610548.3618154}
}

@article{zhang2024layerdiffusion,
  title = {Transparent Image Layer Diffusion Using Latent Transparency},
  author = {Zhang, Lvmin and Agrawala, Maneesh},
  journal = {arXiv preprint},
  year = {2024}
}

@inproceedings{lipman2023flowmatching,
  title = {Flow Matching for Generative Modeling},
  author = {Lipman, Yaron and Chen, Ricky T. Q. and Ben-Hamu, Heli and Nickel, Maximilian and Le, Matt},
  booktitle = {International Conference on Learning Representations},
  year = {2023}
}

@inproceedings{wang2023selfinstruct,
  title = {Self-Instruct: Aligning Language Models with Self-Generated Instructions},
  author = {Wang, Yizhong and Kordi, Yeganeh and Mishra, Swaroop and Liu, Alisa and Smith, Noah A. and Khashabi, Daniel and Hajishirzi, Hannaneh},
  booktitle = {Annual Meeting of the Association for Computational Linguistics},
  year = {2023}
}

@inproceedings{liu2023visualinstruction,
  title = {Visual Instruction Tuning},
  author = {Liu, Haotian and Li, Chunyuan and Wu, Qingyang and Lee, Yong Jae},
  booktitle = {Advances in Neural Information Processing Systems},
  year = {2023}
}

@inproceedings{gadre2023datacomp,
  title = {DataComp: In Search of the Next Generation of Multimodal Datasets},
  author = {Gadre, Samir Yitzhak and Ilharco, Gabriel and Fang, Alex and Hayase, Jonathan and Smyrnis, Georgios and Nguyen, Thao and Marten, Robert and Wortsman, Mitchell and others},
  booktitle = {Advances in Neural Information Processing Systems},
  year = {2023}
}

@book{mayer2009multimedia,
  title = {Multimedia Learning},
  author = {Mayer, Richard E.},
  edition = {2nd},
  publisher = {Cambridge University Press},
  year = {2009}
}

@article{clark1991dual,
  title = {Dual Coding Theory and Education},
  author = {Clark, James M. and Paivio, Allan},
  journal = {Educational Psychology Review},
  volume = {3},
  number = {3},
  pages = {149--210},
  year = {1991}
}

@inproceedings{bateman2010useful,
  title = {Useful Junk? The Effects of Visual Embellishment on Comprehension and Memorability of Charts},
  author = {Bateman, Scott and Mandryk, Regan L. and Gutwin, Carl and Genest, Aaron and McDine, David and Brooks, Christopher},
  booktitle = {Proceedings of the SIGCHI Conference on Human Factors in Computing Systems (CHI)},
  pages = {2573--2582},
  year = {2010}
}

@article{bhat2023infographics,
  title = {Infographics in Educational Settings: A Literature Review},
  author = {Bhat, S. A. and Alyahya, S.},
  journal = {IEEE Access},
  volume = {12},
  pages = {1633--1649},
  year = {2024},
  doi = {10.1109/ACCESS.2023.3348083}
}

@inproceedings{jain2026nerfify,
  title     = {NERFIFY: A Multi-Agent Framework for Turning NeRF Papers into Code},
  author    = {Jain, Seemandhar and Gupta, Keshav and Gupta, Kunal and Chandraker, Manmohan},
  booktitle = {Proceedings of the IEEE/CVF Conference on Computer Vision and Pattern Recognition (CVPR)},
  year      = {2026},
  pages     = {24384--24394}
}

@misc{huang2026scifig,
  title         = {SciFig: Towards Automating Editable Figure Generation for Scientific Papers},
  author        = {Huang, Siyuan and Zhou, Yifan and Gao, Yutong and
                   Yin, Zi and Bai, Juyang and Liu, Xinxin and
                   Chellappa, Rama and Lau, Chun Pong and Peng, Cheng and
                   Nag, Sayan and Pramanick, Shraman},
  year          = {2026},
  eprint        = {2601.04390},
  archivePrefix = {arXiv},
  primaryClass  = {cs.AI}
}

@inproceedings{rodriguez2025starvector,
  title = {StarVector: Generating Scalable Vector Graphics Code from Images and Text},
  author = {Rodriguez, Juan A. and Puri, Abhay and Agarwal, Shubham and Laradji, Issam H. and Rodriguez, Pau and Rajeswar, Sai and Vazquez, David and Pal, Christopher and Pedersoli, Marco},
  booktitle = {IEEE/CVF Conference on Computer Vision and Pattern Recognition (CVPR)},
  year = {2025}
}

@article{wu2023iconshop,
  title = {IconShop: Text-Guided Vector Icon Synthesis with Autoregressive Transformers},
  author = {Wu, Ronghuan and Su, Wanchao and Ma, Kede and Liao, Jing},
  journal = {ACM Transactions on Graphics},
  year = {2023}
}

@article{xing2024llm4svg,
  title = {Empowering LLMs to Understand and Generate Complex Vector Graphics},
  author = {Xing, Ximing and Hu, Juncheng and Liang, Guotao and Zhang, Jing and Xu, Dong and Yu, Qian},
  journal = {arXiv preprint arXiv:2412.11102},
  year = {2024}
}

@article{wu2024chat2svg,
  title = {Chat2SVG: Vector Graphics Generation with Large Language Models and Image Diffusion Models},
  author = {Wu, Ronghuan and Su, Wanchao and Liao, Jing},
  journal = {arXiv preprint arXiv:2411.16602},
  year = {2024}
}

@inproceedings{rombach2022ldm,
  title = {High-Resolution Image Synthesis with Latent Diffusion Models},
  author = {Rombach, Robin and Blattmann, Andreas and Lorenz, Dominik and Esser, Patrick and Ommer, Bj{\"o}rn},
  booktitle = {IEEE/CVF Conference on Computer Vision and Pattern Recognition (CVPR)},
  year = {2022}
}

@inproceedings{saharia2022imagen,
  title = {Photorealistic Text-to-Image Diffusion Models with Deep Language Understanding},
  author = {Saharia, Chitwan and Chan, William and Saxena, Saurabh and Li, Lala and Whang, Jay and Denton, Emily and Ghasemipour, Seyed Kamyar Seyed and Ayan, Burcu Karagol and Mahdavi, S. Sara and Lopes, Rapha Gontijo and Salimans, Tim and Ho, Jonathan and Fleet, David J. and Norouzi, Mohammad},
  booktitle = {Advances in Neural Information Processing Systems (NeurIPS)},
  year = {2022}
}

@inproceedings{podell2023sdxl,
  title = {SDXL: Improving Latent Diffusion Models for High-Resolution Image Synthesis},
  author = {Podell, Dustin and English, Zion and Lacey, Kyle and Blattmann, Andreas and Dockhorn, Tim and M{\"u}ller, Jonas and Penna, Joe and Rombach, Robin},
  booktitle = {International Conference on Learning Representations (ICLR)},
  year = {2024}
}

@article{chen2025diffusionsurvey,
  title = {Comprehensive Exploration of Diffusion Models in Image Generation: A Survey},
  author = {Chen, Hang and Xiang, Qian and Hu, Jiaxin and Ye, Meilin and Yu, Chao and Cheng, Hao and Zhang, Lei},
  journal = {Artificial Intelligence Review},
  volume = {58},
  number = {99},
  year = {2025},
  doi = {10.1007/s10462-025-11110-3}
}

@inproceedings{esser2024sd3,
  title = {Scaling Rectified Flow Transformers for High-Resolution Image Synthesis},
  author = {Esser, Patrick and Kulal, Sumith and Blattmann, Andreas and Entezari, Rahim and M{\"u}ller, Jonas and Saini, Harry and Levi, Yam and Lorenz, Dominik and Sauer, Axel and Boesel, Frederic and Podell, Dustin and Dockhorn, Tim and English, Zion and Lacey, Kyle and Goodwin, Alex and Marek, Yannik and Rombach, Robin},
  booktitle = {International Conference on Machine Learning (ICML)},
  year = {2024}
}

@inproceedings{liu2023characteraware,
  title = {Character-Aware Models Improve Visual Text Rendering},
  author = {Liu, Rosanne and Garrette, Dan and Saharia, Chitwan and Chan, William and Roberts, Adam and Narang, Sharan and Blok, Irina and Mical, RJ and Norouzi, Mohammad and Constant, Noah},
  booktitle = {Annual Meeting of the Association for Computational Linguistics (ACL)},
  year = {2023}
}

@misc{qwen2025image2512,
  author = {{Qwen Team}},
  title = {{Qwen-Image-2512}},
  year = {2025},
  howpublished = {\url{https://qwen.ai/blog?id=qwen-image-2512}},
  note = {Technical blog}
}

@misc{qwen2025imageedit,
  author = {{Qwen Team}},
  title = {{Qwen-Image-Edit}: Image Editing with Higher Quality and Efficiency},
  year = {2025},
  howpublished = {\url{https://qwen.ai/blog?id=qwen-image-edit}},
  note = {Technical blog}
}

@software{gatis2025rembg,
  author = {Gatis, Daniel},
  title = {rembg: A Tool to Remove Image Backgrounds},
  year = {2025},
  version = {2.0.66},
  url = {https://github.com/danielgatis/rembg},
  note = {MIT-licensed software}
}

@manual{lachowicz2014rsvgconvert,
  author = {Lachowicz, Dom and Moore, Caleb and others},
  title = {{rsvg-convert}: Turn {SVG} Files into Raster Images},
  organization = {librsvg},
  year = {2014},
  howpublished = {Ubuntu 14.04 manual page},
  url = {https://manpages.ubuntu.com/manpages/trusty/man1/rsvg-convert.1.html}
}

@article{van_der_Walt_2014,
   title={scikit-image: image processing in Python},
   volume={2},
   ISSN={2167-8359},
   url={http://dx.doi.org/10.7717/peerj.453},
   DOI={10.7717/peerj.453},
   journal={PeerJ},
   publisher={PeerJ},
   author={van der Walt, Stéfan and Schönberger, Johannes L. and Nunez-Iglesias, Juan and Boulogne, François and Warner, Joshua D. and Yager, Neil and Gouillart, Emmanuelle and Yu, Tony},
   year={2014},
   month=June, pages={e453} }

@misc{tesseractocr2026,
  author = {{Tesseract OCR contributors}},
  title = {{Tesseract}: Open Source {OCR} Engine},
  year = {2026},
  howpublished = {\url{https://github.com/tesseract-ocr/tesseract}},
  note = {Software repository; accessed September 2, 2026}
}

@misc{ideogram2026v4,
  author = {{Ideogram AI}},
  title = {{Ideogram 4.0}},
  year = {2026},
  howpublished = {\url{https://ideogram.ai/blog/ideogram-4.0/}},
  note = {Technical blog}
}

@misc{anthropic2026claudeopus48,
  author = {{Anthropic}},
  title = {Introducing {Claude Opus 4.8}},
  year = {2026},
  month = may,
  howpublished = {\url{https://www.anthropic.com/news/claude-opus-4-8}},
  note = {Product announcement, May 28, 2026}
}

@misc{openai2026gpt56sol,
  author = {{OpenAI}},
  title = {{GPT-5.6 Sol} Model},
  year = {2026},
  howpublished = {\url{https://developers.openai.com/api/docs/models/gpt-5.6-sol}},
  note = {Model documentation}
}

@misc{deepmind2026gemini37flash,
  author = {{Google DeepMind}},
  title = {{Gemini 3.7 Flash} Model Card},
  year = {2026},
  month = aug,
  howpublished = {\url{https://deepmind.google/models/model-cards/gemini-3-7-flash/}},
  note = {Published August 13, 2026}
}

@misc{w3cSvg11,
  author = {{W3C SVG Working Group}},
  title = {{Scalable Vector Graphics (SVG) 1.1 (Second Edition)}},
  year = {2011},
  howpublished = {W3C Recommendation},
  url = {https://www.w3.org/TR/SVG11/},
}

@article{kingma2013auto,
  title = {Auto-Encoding Variational Bayes},
  author = {Kingma, Diederik P. and Welling, Max},
  journal = {arXiv preprint arXiv:1312.6114},
  year = {2013}
}

@article{yang2024layerdecomp,
  title         = {Generative Image Layer Decomposition with Visual Effects},
  author        = {Yang, Jinrui and Liu, Qing and Li, Yijun and Kim, Soo Ye
                   and Pakhomov, Daniil and Ren, Mengwei and Zhang, Jianming
                   and Lin, Zhe and Xie, Cihang and Zhou, Yuyin},
  year          = {2024},
  eprint        = {2411.17864},
  archivePrefix = {arXiv},
  primaryClass  = {cs.CV},
  url           = {https://arxiv.org/abs/2411.17864}
}

@article{yin2025qwenimagelayered,
  title         = {{Qwen-Image-Layered}: Towards Inherent Editability via
                   Layer Decomposition},
  author        = {Yin, Shengming and Zhang, Zekai and Tang, Zecheng
                   and Gao, Kaiyuan and Xu, Xiao and Yan, Kun and Li, Jiahao
                   and Chen, Yilei and Chen, Yuxiang and Shum, Heung-Yeung
                   and Ni, Lionel M. and Zhou, Jingren and Lin, Junyang
                   and Wu, Chenfei},
  year          = {2025},
  eprint        = {2512.15603},
  archivePrefix = {arXiv},
  primaryClass  = {cs.CV},
  url           = {https://arxiv.org/abs/2512.15603}
}

@article{wang2026reveallayer,
  title         = {{RevealLayer}: Disentangling Hidden and Visible Layers via
                   Occlusion-Aware Image Decomposition},
  author        = {Wang, Binhao and Zhao, Shihao and Cheng, Bo and Ji, Qiuyu
                   and Ma, Yuhang and Wu, Liebucha and Liu, Shanyuan
                   and Leng, Dawei and Yin, Yuhui},
  year          = {2026},
  eprint        = {2605.11818},
  archivePrefix = {arXiv},
  primaryClass  = {cs.CV},
  url           = {https://arxiv.org/abs/2605.11818}
}
\bibliographystyle{iclr2027_conference}

\clearpage
\appendix
\begin{center}
  {\LARGE\bfseries Appendix\par}
\end{center}

\vspace{1em}

\subsection*{Contents}

{\small
\begin{itemize}
  \setlength{\itemsep}{0.35em}
  \item \hyperref[app:data]{Appendix~\ref*{app:data}: Data}
    \begin{itemize}
      \setlength{\itemsep}{0.1em}
      \item \hyperref[app:data-sourcing]{\ref*{app:data-sourcing} Concept sourcing and topical scope}
      \item \hyperref[app:data-generation]{\ref*{app:data-generation} Multi-agent generation pipeline}
      \item \hyperref[app:data-validation]{\ref*{app:data-validation} Layer decomposition, augmentation, and validation}
      \item \hyperref[app:data-statistics]{\ref*{app:data-statistics} Dataset composition, splits, and provenance}
      \item \hyperref[app:data-analysis]{\ref*{app:data-analysis} Dataset analysis: color composition is coupled and learnable}
      \item \hyperref[app:data-implementation]{\ref*{app:data-implementation} Implementation details}
    \end{itemize}
  \item \hyperref[app:additional-experiments]{Appendix~\ref*{app:additional-experiments}: Additional Details and Experiments}
    \begin{itemize}
      \setlength{\itemsep}{0.1em}
      \item \hyperref[app:method-details]{\ref*{app:method-details} Additional method details}
      \item \hyperref[app:evaluation-details]{\ref*{app:evaluation-details} Evaluation details}
      \item \hyperref[app:additional-comparisons]{\ref*{app:additional-comparisons} Additional comparisons}
      % \item \hyperref[app:transformations]{\ref*{app:transformations} Controlled transformations and edit locality}
      \item \hyperref[app:systems-profile]{\ref*{app:systems-profile} Systems profile}
    \end{itemize}
  \item \hyperref[app:downstream-applications]{Appendix~\ref*{app:downstream-applications}: Downstream Applications}
    \begin{itemize}
      \setlength{\itemsep}{0.1em}
      \item \hyperref[app:downstream-text-editing-localization]{\ref*{app:downstream-text-editing-localization} Text editing and localization}
      \item \hyperref[app:downstream-animation-progressive-reveal]{\ref*{app:downstream-animation-progressive-reveal} Animation and progressive reveal}
      \item \hyperref[app:downstream-asset-reuse-recomposition]{\ref*{app:downstream-asset-reuse-recomposition} Asset reuse and cross-document recomposition}
      \item \hyperref[app:downstream-layout-retargeting]{\ref*{app:downstream-layout-retargeting} Layout retargeting and responsive resize}
      % \item \hyperref[app:downstream-style-retheming]{\ref*{app:downstream-style-retheming} Style transfer and re-theming}
      \item \hyperref[app:downstream-command-driven-editing]{\ref*{app:downstream-command-driven-editing} Command-driven interactive editing}
      \item \hyperref[app:downstream-svg-refinement-propagation]{\ref*{app:downstream-svg-refinement-propagation} SVG-space refinement and faithful diffusion propagation}
    \end{itemize}
  \item \hyperref[app:limitations]{Appendix~\ref*{app:limitations}: Limitations, ethics, and broader impact}
\end{itemize}
}

% \clearpage
\section{Data}
\label{app:data}

Training \method{} requires a form of supervision that, to our knowledge, no
existing corpus provides: infographics paired at the level of individual,
renderer-aligned layers. A usable training record must expose, for a single
graphic, (i) a structural program that fixes geometry, hierarchy, paint order,
and typography; (ii) an aestheticized, text-free rendering of that program; and
(iii) a decomposition of that rendering into ordered, mutually non-overlapping
RGBA layers with amodal and visible masks. Public SVG datasets supply
programs but no rich appearance, while raster infographic collections supply
appearance but discard element ownership, occlusion, and text metadata. We
therefore \emph{synthesize} the corpus with a multi-agent generation pipeline
that couples a large language model, which designs and validates editable SVG
programs, with a diffusion image editor, which aestheticizes only the non-text
visual content. The pipeline is designed so that every artifact \method{}
consumes at training time, namely the colorless structural input, the text-free
target, the per-layer amodal/visible assets, and the graph of overlap and
hierarchy relations, is produced deterministically from an SVG the agent has
already checked, rather than recovered from pixels after the fact.

This section describes how concepts are sourced from an educational course
catalog (\S\ref{app:data-sourcing}), how the multi-agent pipeline turns each
concept into a validated paired record (\S\ref{app:data-generation}), how those
records are decomposed into layers, augmented, and validated
(\S\ref{app:data-validation}), the resulting scale, splits, and provenance
(\S\ref{app:data-statistics}), an analysis of the appearance statistics the
pipeline induces (\S\ref{app:data-analysis}), and the implementation and compute
details needed to reproduce it (\S\ref{app:data-implementation}).

\begin{figure}[t]
  \centering
  \includegraphics[width=\linewidth]{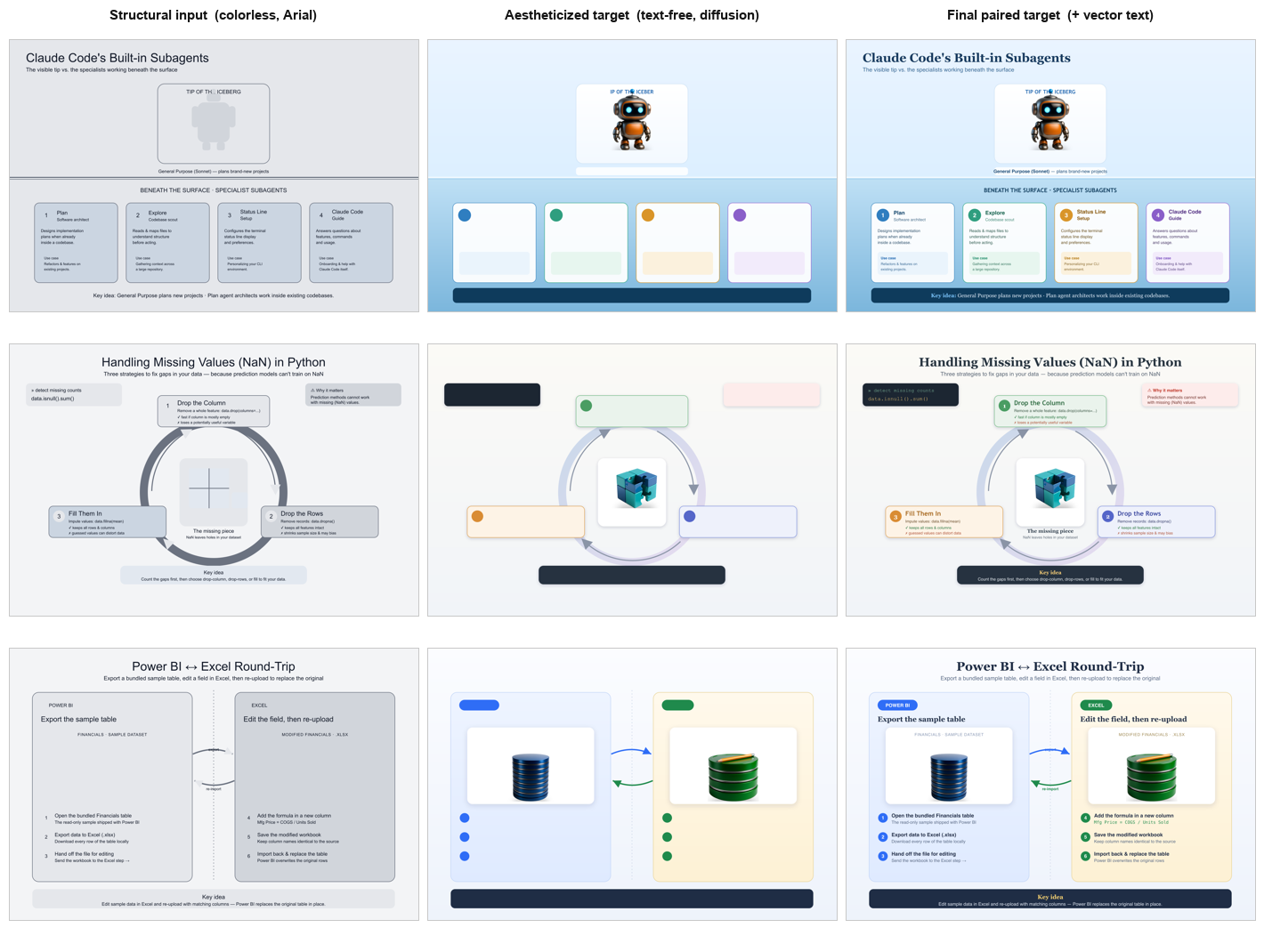}
  \caption{\textbf{The paired training record.} Each column is one artifact the
  pipeline emits for the same sample. \emph{Left:} the structural input, an SVG
  with color, fonts, and textures stripped and text set in a neutral face, which
  supplies geometry, hierarchy, and paint order without leaking target
  appearance. \emph{Middle:} the aestheticized, \emph{text-free} target that the
  diffusion streams are trained to produce; the editable-visual regions have been
  restyled into rich illustrations while every text region is left blank.
  \emph{Right:} the final paired target after the preserved vector text layer is
  composited on top. Diffusion is never asked to render glyphs.}
  \label{fig:data-pair}
\end{figure}

\subsection{Concept sourcing and topical scope}
\label{app:data-sourcing}

We seed generation with concepts drawn from popular online educational
courses. Each concept is a short, self-contained instructional idea
(for example, ``handling missing values in Python,'' ``the tip-of-the-iceberg
model of specialist agents,'' or ``the Power BI\,$\leftrightarrow$\,Excel
round-trip''), and it carries the identifier and title of the course it was
extracted from. Anchoring the corpus in real course material keeps the
distribution of topics, terminology, and visual conventions close to genuine
educational infographics (dense labels, step-wise procedures, comparisons,
cycles, and metaphorical scenes) rather than to generic stock imagery. The
course identifier is retained on every derived record and is used both for
provenance and, critically, for constructing leakage-free splits
(\S\ref{app:data-statistics}).

\subsection{Multi-agent generation pipeline}
\label{app:data-generation}

The pipeline proceeds in four stages. A design stage authors a structured,
editable SVG program from a concept; an extraction stage renders the structural
input and the text-free layout and serializes structured conditioning; an
aestheticization stage restyles each editable region with a diffusion image
editor; and a composition stage mattes and reinserts the styled illustrations in
paint order. All appearance in the final target originates in the SVG program or
in the diffusion edits of its tagged regions, never in a free-form raster of the
whole scene.

\paragraph{Stage A: structured SVG design (LLM).}
A design agent, built on a large language model, turns a concept into a
complete, standalone SVG on a fixed $1536\times1024$ canvas. Beyond producing an
aesthetic educational layout, the agent must mark at least one
\emph{editable-visual} group: a coarse vector placeholder standing in for a
tangible, photographable subject (a robot, a database cylinder, a brass scale, a
stack of coins). Each such group is emitted with a strict, machine-parseable
signature,
\begin{center}
\texttt{\textless g id="editable-visual-\textit{k}" data-role="editable-visual"}\\
\texttt{~~~~data-visual-desc="\dots" data-edit-prompt="\dots"\textgreater\dots\textless/g\textgreater},
\end{center}
where \texttt{data-edit-prompt} is the exact, self-contained instruction later
handed to the diffusion editor. To prevent the corpus from collapsing onto a
single visual template, the agent is assigned one of fourteen layout archetypes
per concept (hero-left, hub-and-spoke, comparison-split, circular-cycle,
above-below, quadrant-grid, timeline, and others), rotated deterministically
across a batch and treated as an adaptable starting structure rather than a rigid
template. The prompt enforces explicit layout discipline (shared alignment
gridlines, equal gaps for repeated elements, symmetry matched to the number of
editable visuals) and an aesthetic direction tuned for legibility (light airy
backgrounds, per-section color coding, a strong title/subtitle/takeaway
hierarchy). A guard explicitly forbids screens, dashboards, spreadsheets, and
other flat, text-bearing objects from being marked editable, because such
subjects photorealize poorly and reintroduce garbled text; those elements remain
crisp vector content. The agent operates in a render-and-inspect loop: it
rasterizes its SVG, examines the image, and performs at least one revision pass
to correct misalignment, uneven spacing, dead space, or low-contrast text before
emitting a manifest of its editable groups.

\paragraph{Stage B: structural extraction and conditioning.}
From the validated SVG we deterministically derive the model's inputs. The
\emph{structural input} (Fig.~\ref{fig:data-pair}, left) is produced by
``rawifying'' the SVG, that is, stripping color, fonts, gradients, and textures,
setting text in a neutral face, and rendering to a flat raster, so that it encodes
placement, grouping, and paint order without revealing target appearance. We
also render a \emph{text-free layout} in which all recursive text nodes are
removed. The record captures the scene as typed metadata: every text run with its content, semantic
role (title/heading/label), normalized bounding box, font family, size, weight,
and fill; every illustration with its identifier, description, the
\texttt{data-edit-prompt}, its bounding box, and its z-index; and a global style
block listing the design palette (hex swatches), font families, and mean corner
radius. This record is the interface between deterministic structure and
diffusion aestheticization used throughout the method.

\paragraph{Stage C: illustration aestheticization (diffusion editor).}
Each editable-visual crop is rendered in isolation and passed to a diffusion
image editor (Qwen-Image-Edit \cite{qwen2025imageedit}) that repaints \emph{only} that region into a rich
illustration on a plain white studio background. A style controller makes the
requested look dominant: it strips residual realism adjectives from the agent's
subject description, prepends a leading style directive, and swaps a matching
negative prompt, so a single subject can be realized in any of six styles
(\textit{photoreal}, \textit{flat2d}, \textit{soft3d}, \textit{cartoon},
\textit{watercolor}, \textit{isometric}) while its identity and composition are
preserved. The middle column of Fig.~\ref{fig:data-pair} shows the resulting
text-free targets, in which the coarse placeholders have become detailed objects
but every label region remains blank.

\paragraph{Stage D: matting and z-ordered composition.}
The styled illustration is cut from its white background with an alpha matte
(rembg/u2net~\citep{gatis2025rembg}), cleaned to its largest connected component and feathered, fitted
to the placeholder's bounding box, and composited back into the scene at exactly
the illustration's original z-index. Because the surrounding layout is rebuilt
from the SVG with the editable groups removed, the realistic render occupies the
placeholder footprint without disturbing any text or vector element, and soft
contact shadows are added programmatically and kept off text and table regions.
The output is the final paired target (Fig.~\ref{fig:data-pair}, right): the
aestheticized, text-free composite with the preserved vector text layer rendered
crisply on top.

\subsection{Layer decomposition, augmentation, and validation}
\label{app:data-validation}

\paragraph{Ordered non-overlap layer decomposition.}
For every sample we additionally emit the amodal layer decomposition. Each top-level layer is rendered in isolation
to recover its \emph{amodal} silhouette (its complete support as if nothing
occluded it) and its \emph{visible} mask (the pixels surviving occlusion by all
higher-z layers); their difference is exactly the occluded region. Editable
layers take their amodal pixels from the matted realistic render, reconstructed
with the same parameters used in composition, so that they agree with the final
target. We also record, for every pair of layers whose amodal silhouettes
intersect, an overlap edge annotated with the back/front assignment and the
covered fraction. These masks and edges are what make the ordered non-overlap
layer sets and the graph-biased communication of the method well defined; they
are computed deterministically from the renderer, not discovered from pixels.

\paragraph{Style augmentation via illustration reuse.}
Because Stage A (LLM design) and Stage B (structural extraction) are the
expensive, language-model-dependent parts of the pipeline, we amortize them: given an
already-built sample with its per-layer amodal RGBA and its tagged
editable-visual illustration, we generate additional $(\text{input}\to
\text{target})$ pairs whose only difference is the visual \emph{style} of the
illustration. The existing target layers and structured conditioning are reused
verbatim; only Stage C (re-style the illustration) and Stage D (re-matte,
re-compose) are rerun on-GPU. This yields multiple stylistic realizations of the
same structural layout at a fraction of the cost, and a denser-composition
variant reuses previously generated illustrations inside new, higher-overlap
layouts to increase occlusion depth.

\paragraph{Validation and quality control.}
Records are filtered at several points. The design agent's own inspect-and-revise
loop rejects layouts with misalignment, empty regions, or unreadable text before
a sample is emitted. Structural extraction validates that text nodes are fully
removed from visual targets and that every illustration crop is isolated with
clear margins. Composition validates alpha coverage of the matte and geometric
fit within the target footprint, and layer reconstruction is checked against the
composite so that within-group elements do not overwrite one another and every
cross-group overlap follows SVG paint order. Samples failing these checks are
regenerated or dropped.

\subsection{Dataset composition, splits, and provenance}
\label{app:data-statistics}

\begin{figure}[t]
  \centering
  \includegraphics[width=\linewidth]{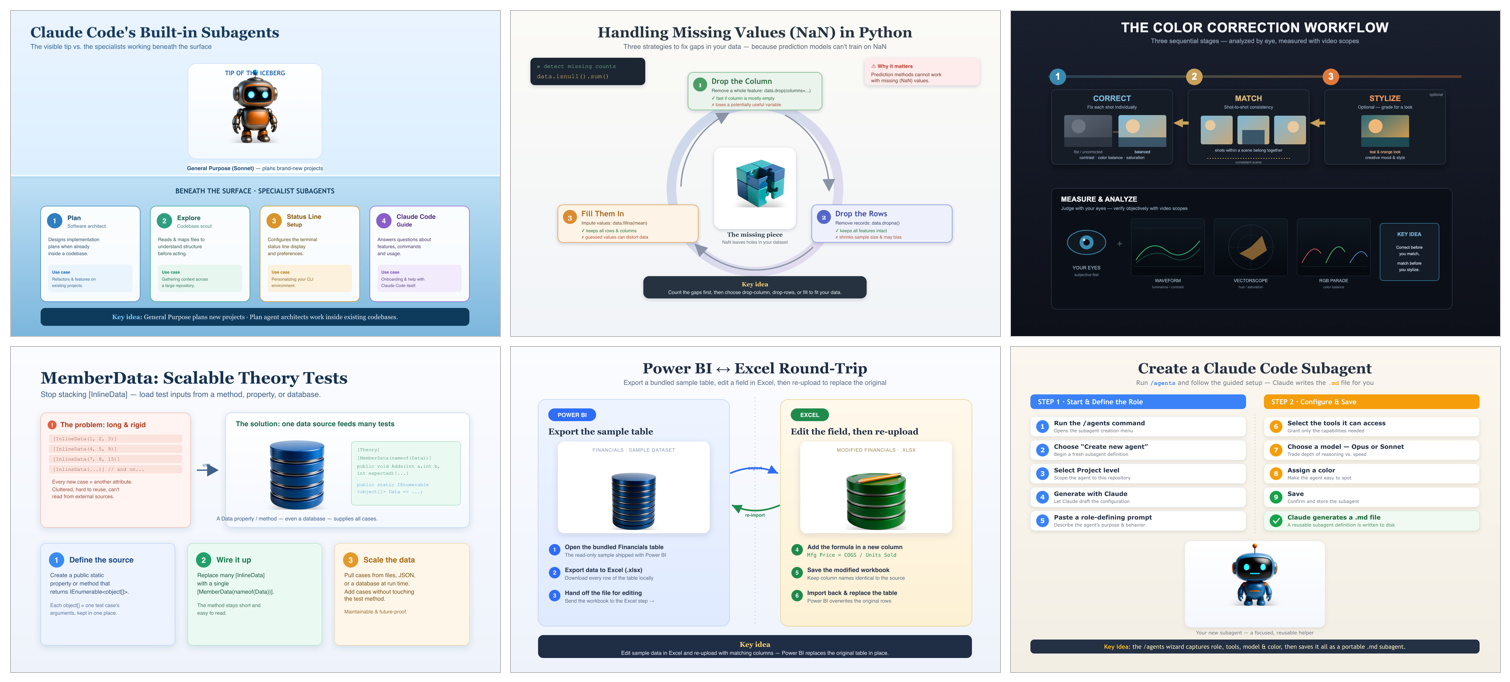}
  \caption{\textbf{Representative dataset infographics.} The corpus spans diverse
  layout archetypes (above-below iceberg, circular cycle, hero-plus-steps,
  comparison split, guided procedure, staged timeline), both light and dark
  themes, and a range of illustration styles, while preserving crisp typography
  and clean editable-visual footprints in every sample. Text is always rendered
  from vector metadata, never by the diffusion editor.}
  \label{fig:data-gallery}
\end{figure}

The merged corpus contains $23{,}960$ paired records, combining an initial
generation pass (roughly $8.4$K pairs) with the style-augmentation and
dense-reuse passes (roughly $15.5$K pairs); the passes are merged so that no
structural layout appears across the boundary in conflicting roles. Every record
carries the full training tuple: the structural input, the text-free and final
targets, the per-layer amodal/visible masks, the overlap and hierarchy graph,
and the structured conditioning metadata. Illustration styles are distributed roughly
evenly across the six style families, and the samples span the fourteen layout
archetypes. Figure~\ref{fig:data-gallery} shows representative final targets.

\paragraph{Leakage-free, course-disjoint splits.}
Splits are constructed at the granularity of the \emph{source course}, not the
individual sample, so that no concept from a course used for evaluation can also
appear in training. Courses are ordered by a fixed hash of their identifier, and
a subset-sum procedure selects a set of whole courses summing to $200$ samples
for the test split and, from the remainder, another $200$ for validation; all
remaining courses form the training split. Because a course contributes all of
its concepts to exactly one split, the evaluation sets measure generalization to
unseen courses and topics rather than memorization of a familiar course's visual
style.

\paragraph{Provenance.}
Each record retains its course identifier and title, the assigned layout
archetype, the editable-visual descriptions and edit prompts, the diffusion
style and seed, and the compositing parameters, so that any sample can be traced
back to its concept and regenerated. Concepts are used as textual seeds for
synthesis; the visual content of every target is generated by our own pipeline.

\subsection{Dataset analysis: color composition is coupled and learnable}
\label{app:data-analysis}

A natural question for a synthesized corpus is whether it contains
\emph{learnable structure} beyond each element in isolation, in particular,
whether the colors of different elements within a graphic are related, as they
are in human-designed infographics where palette choices are made jointly. We
quantify this on a $1{,}000$-sample subset by comparing, within each graphic,
the colors of the vector design and the colors of the diffusion-generated
illustrations, all in perceptual CIELAB space.

\begin{figure}[t]
  \centering
  \includegraphics[width=\linewidth]{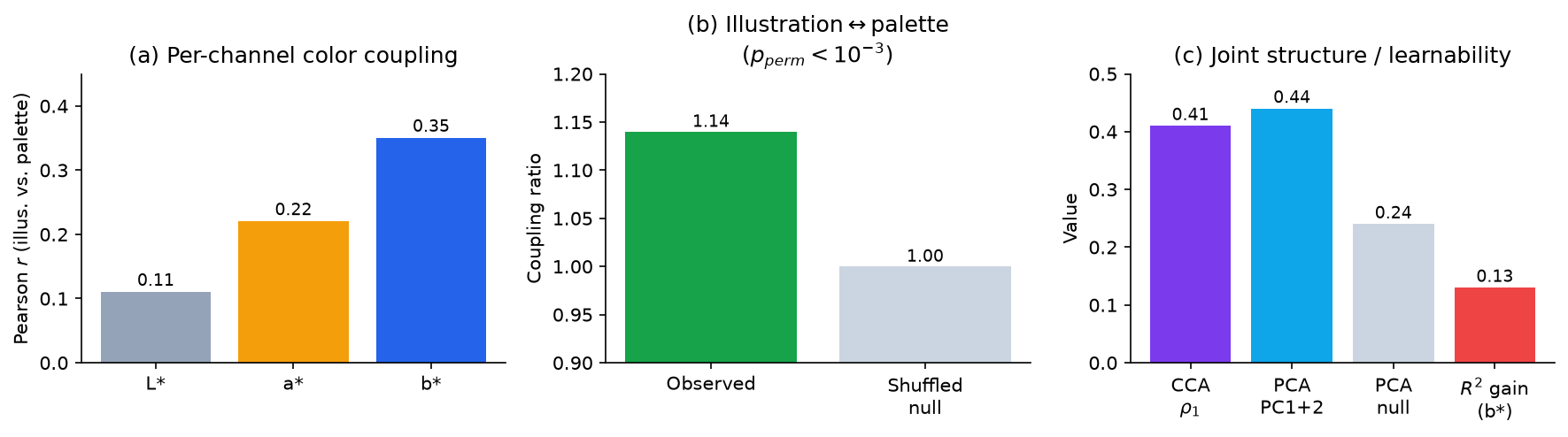}
  \caption{\textbf{Cross-element color structure ($n=1{,}000$).}
  (a) Per-channel Pearson correlation in CIELAB between mean illustration color
  and mean palette color; coupling is strongest on the $b^\ast$ (blue-yellow)
  axis. (b) Observed illustration$\leftrightarrow$palette coupling exceeds a
  shuffled null ($p_{\mathrm{perm}}<10^{-3}$). (c) Joint-structure and
  learnability summaries: canonical correlation $\rho_1$, variance explained by
  the first two joint principal components versus a null, and the $R^2$ gain on
  $b^\ast$ from conditioning on the palette. Together these show the dataset
  induces coordinated, non-random color composition across layers.}
  \label{fig:color-corr}
\end{figure}

Two findings emerge (Fig.~\ref{fig:color-corr}). First, the vector side is
palette-coherent by construction: text fills are drawn \emph{verbatim} from the
design palette (median $\Delta E=0$ to the nearest palette swatch), confirming
that the LLM designs a single coherent color scheme per graphic. Second, and less
trivially, the \emph{diffusion-generated illustrations are measurably coupled to
that vector palette} even though the editor sees only a per-illustration subject
prompt. Aggregate illustration color and palette color are positively correlated
per channel, most strongly on the blue-yellow ($b^\ast$, $r=0.35$) and
red-green ($a^\ast$, $r=0.22$) axes and weakest on lightness ($L^\ast$,
$r=0.11$); a permutation test rejects independence ($p_{\mathrm{perm}}<10^{-3}$,
observed-to-null coupling ratio $1.14\times$); and a canonical correlation
analysis between illustration and palette color reaches a first correlation of
$\rho_1=0.41$. The coupling is also \emph{learnable}: knowing the palette
improves prediction of an illustration's $b^\ast$ by $\Delta R^2=+0.13$ beyond
the style label alone, and the leading two principal components of the joint
color representation explain $0.44$ of the variance versus $0.24$ under a
shuffled null.

Mechanistically, this coupling is \emph{authorial} rather than injected: the same
language model writes both the design palette and each illustration's subject
prompt, so color intent leaks through word choice, and it survives the editor's
independent re-styling. This is a desirable property for our setting: it means
the corpus rewards a model that coordinates appearance \emph{across} layers
rather than styling each layer in isolation, which is precisely the behavior the
correlated, communicating streams of \method{} are designed to capture.

\subsection{Implementation details}
\label{app:data-implementation}

The design agent is driven by a large language model (Claude Opus 4.8 \cite{anthropic2026claudeopus48}), accessed
through a standard chat-completion API, and invoked with a render-and-inspect
tool loop that supplies the rasterized SVG back to the model for revision. SVGs
are rendered with \texttt{rsvg-convert}~\citep{lachowicz2014rsvgconvert} at the fixed $1536\times1024$ canvas.
Illustration aestheticization uses the Qwen-Image-Edit \cite{qwen2025imageedit} diffusion image editor in
\texttt{bfloat16}, with classifier-free guidance $5.0$, $40$ denoising steps, and
a fixed seed for reproducibility; crop sizes are snapped to multiples of $32$
within $[256,1024]$. Standalone asset generation, where needed, uses the
Qwen-Image \cite{qwen2025image2512} text-to-image model. Background removal uses rembg \cite{gatis2025rembg} with the u2net
matting model; a border-connected white-key matte is available as a fast
alternative for saturated subjects. Color analysis is performed in CIELAB (via
\texttt{scikit-image}~\citep{van_der_Walt_2014}) with $k$-means dominant-color extraction and permutation,
canonical-correlation, and PCA baselines.

The stages have different resource needs, so we separate them. The
language-model stage requires only API access and runs on a CPU host, while the
GPU stages (aestheticization, matting, composition, and layer rendering) run on
NVIDIA H100 and H200 GPUs. Large batches are executed on 8$\times$H100 and
8$\times$H200 nodes. The
style-augmentation and dense-reuse passes require no language-model access, since
they reuse the previously validated structural layers and only rerun the GPU
restyle-and-compose steps.

\clearpage
\section{Additional Details and Experiments}
\label{app:additional-experiments}

This section provides the complete method specification followed by
supplementary experimental details, comparisons, and analyses.

\subsection{Additional method details}
\label{app:method-details}

\paragraph{Renderer-derived masks and hidden areas.}
The offline renderer removes all SVG text nodes and records a text-free
target, per-element RGBA assets, element types, normalized boxes,
parent-child links, paint order, and masks. For element \(e_i\), the full
(amodal) mask \(A_i\) covers its entire area before later elements are
painted, while \(V_i\) retains
only the pixels visible in the final rendering. Their difference
\(H_i=(A_i-V_i)_+\) identifies content hidden by an element in front.
Because
per-element assets are rendered before composition, \(H_i\) contains the
element's true appearance rather than a blank region. This provides a direct
training target for content revealed by deletion, translation, or z-order edits.
Illustration prompts and text regions are routed to their owning visual
elements, although text remains absent from image targets.

\paragraph{Properties of ordered grouping.}
The rule in Eq.~\eqref{eq:ordered-coloring} never places an element in the
same or an earlier group than an already processed element that it overlaps.
Thus every overlap keeps SVG paint order, while
Eq.~\eqref{eq:nonoverlap} guarantees that elements sharing a generated layer cannot
overwrite one another. The groups are not based on meaning or element type:
different element types may share a generated layer whenever their full masks do not
overlap. The layers are also not generated one at a time; the group index
determines composition order but all layers are sampled simultaneously.
After a
geometric edit, grouping and graph information are recomputed without resampling
unaffected layers.

\paragraph{Correlated layer-wise flow matching.}
Let \(x_{1,g}\in\mathbb{R}^{K\times D}\) be the packed internal VAE features
of the clean target for group \(g\). We combine shared and private Gaussian noise,
\begin{equation}
 \epsilon_g=\sqrt{\rho}\,\epsilon_{\rm sh}
              +\sqrt{1-\rho}\,\epsilon_g^{\rm pr},
 \quad
 \epsilon_{\rm sh},\epsilon_g^{\rm pr}\sim\mathcal{N}(0,I),
 \label{eq:noise}
\end{equation}
where the shared component encourages a consistent scene-level look and the
private component preserves layer-specific variation. The current schedule samples
\(\rho\sim\mathcal{U}(0.5,1)\). Each example uses one common timestep
\(t\), so communication always relates layers at the same noise level.
The implemented linear path is
\begin{equation}
 x_{t,g}=(1-t)\epsilon_g+t x_{1,g},\qquad
 u_g=x_{1,g}-\epsilon_g,\qquad t\sim\mathcal{U}(0,1),
 \label{eq:path}
\end{equation}
with primary objective
\begin{equation}
 \mathcal{L}_{\rm layer}
 =\frac{\sum_g m_g\|v_{\theta,g}-u_g\|_2^2}
        {\sum_g m_g},
 \qquad
 \widehat x_{1,g}=x_{t,g}+(1-t)v_{\theta,g},
 \label{eq:flow}
\end{equation}
where \(m_g\) ignores empty layer slots added for batching. During generation, these partly shared
starting states are updated in parallel. Shared noise only helps align the
overall look:
without the communication block, the frozen backbone still processes each
group independently and cannot react to another layer's evolving content or
boundary.
Figure~\ref{fig:layer-diffusion-loop} shows the full noise-to-layer path and
expands one repeated denoising step.

\paragraph{Communication summaries and graph bias.}
At a selected backbone depth, each layer is summarized using nine compact
feature groups: summary statistics from the full and hidden areas, averages
from visible areas and boundary bands, \(2\times2\) pooled spatial context,
type, box, paint order, timestep, local text context, and a global feature
average. These values are combined into the group summary \(h_g\). The graph
tensor \(E_{gh}\in\mathbb{R}^{4}\) stores directed
back-to-front overlap, front-to-back overlap, parent-to-child links, and
child-to-parent links. A learned linear map produces one added bias
per attention head, as formalized in Eq.~\eqref{eq:graph-attention}; invalid
padded groups are masked from both keys and outputs.

\paragraph{Style and boundary paths.}
The communicated summary \(\widetilde h_g\) is returned to the image features
through the two paths in Eq.~\eqref{eq:branches}. The style path predicts a
coarse \(2\times2\) cell grid and enlarges it to provide smooth
appearance corrections. The boundary path combines local hidden states with
\(\widetilde h_g\) and is multiplied by a narrow band \(B_g\), limiting
its effect to silhouettes and cross-layer contacts. Their learned gate biases
are initialized to \([-2.0,-2.5]\), giving initial gate strengths of
approximately \(0.12\) and \(0.076\). These gates start small but not at
zero. The block makes no change at the start of training because the style
and boundary output layers start at zero.

% \paragraph{Adapter configuration and current approximation.}
% The evaluated configuration uses an internal feature size of 128, four
% attention heads, at most 32 groups, and communication at transformer blocks
% 9 and 19. The trainable adapter contains 5,730,596 parameters:
% 1,227,008 in the grouped layer-control adapter and 2,251,794 in each communication
% block. The image backbone remains frozen. Training operates on packed
% internal RGB representations obtained by flattening RGBA targets against a fixed light
% background, while masks provide alpha training targets. During the current
% evaluation stage, decoded layer appearance is combined using
% full-resolution renderer-derived alpha masks; this isolates appearance
% quality from learned mask quality and is treated as a current approximation,
% not as a required part of the method.

\paragraph{Composition and structure losses.}
The loss for each layer does not by itself check the result after the layers
are combined. The composition loss sorts groups by paint order
and compares their masked reconstruction with the clean composite.
\(\mathcal{L}_{\rm hid}\) supervises the hidden area
\(H_g=(A_g-V_g)_+\), and \(\mathcal{L}_{\rm bnd}\) matches values and spatial
gradients near ancestor boundaries. For every auxiliary term, we first recover the one-step clean
estimate \(\widehat x_{1,g}\) for each group from its noisy state and predicted
velocity using Eq.~\eqref{eq:flow}. This estimate remains in the packed model
representation; training does not decode a full RGB image through the VAE for
these losses. Only the composition loss combines the estimated groups: it
applies their amodal masks, sorts them by SVG paint order, and compares the
result with the corresponding composition of the clean target groups. The
remaining auxiliary losses operate directly on individual clean group
estimates.

% A cosine descriptor loss
% \(\mathcal{L}_{\rm id}\), contrastive assignment, pairwise style consistency,
% and a duplicate margin discourage ownership swaps, inconsistent appearance,
% and repeated illustrations. 

The full objective is
\begin{equation}
 \mathcal{L}=\lambda_{\rm layer}\mathcal{L}_{\rm layer}
 +\lambda_{\rm comp}\mathcal{L}_{\rm comp}
 +\lambda_{\rm hid}\mathcal{L}_{\rm hid}
 +\lambda_{\rm bnd}\mathcal{L}_{\rm bnd}
%  +\lambda_{\rm id}\mathcal{L}_{\rm id}
%  +\lambda_{\rm asn}\mathcal{L}_{\rm asn}
%  +\lambda_{\rm sty}\mathcal{L}_{\rm sty}
%  +\lambda_{\rm dup}\mathcal{L}_{\rm dup}.
 \label{eq:total-loss}
\end{equation}
% The layer term receives the largest baseline weight.
%  Auxiliary terms and
% weights are evaluated through ablations.

\paragraph{Exact typography.}
All SVG text nodes are removed from diffusion targets, and prompts
request blank title, label, and caption regions. We retain the Unicode string,
role, bounding box, font family and style, alignment, transforms, hierarchy,
and owning visual element. During generation, the text-free composite is
placed beneath a transparent vector-rendered text layer
\(R_{\rm text}(\mathcal{T})\), as specified in Eq.~\eqref{eq:artifact}.
Spelling, line breaks, alignment, and font edits therefore require only
vector re-rendering, not another diffusion pass.

\subsection{Evaluation details}
\label{app:evaluation-details}

\paragraph{Compared methods.}
The primary comparison (Table~\ref{tab:main}) evaluates four methods:
(i)~\textbf{SVG (LLM-generated)}, a reference in which an LLM (Claude Opus~4.8\cite{anthropic2026claudeopus48})
recolors and completes the structural SVG, which is then rasterized with
\texttt{rsvg-convert\cite{lachowicz2014rsvgconvert}}; this yields exact vector text by construction but is not
a photorealistic synthesizer;
(ii)~\textbf{Qwen-Image-2512}~\citep{qwen2025image2512}, a prompt-only diffusion model conditioned on the
design-concept prompt which is itself derived from the same SVG layout (for fairness in evaluation);
(iii)~\textbf{Ideogram-4}~\citep{ideogram2026v4}, a prompt-only open-weights system whose prompt is
constructed from the same SVG layout, with its structural elements serialized
into the structured prompt format the model expects; and
(iv)~\textbf{\method{}}, grouped layer-set diffusion
followed by exact SVG text compositing.
Prompt-only baselines receive the concept prompt reconstructed from each
record's structured conditioning; \method{} additionally
consumes the structural layer set.

% \paragraph{Held-out set construction.}
% We generated all four methods for 262 unique validation layouts.  The reported
% comparison uses the 137 samples for which a photorealistic variant is available
% across all three raster baselines (Qwen-Image-2512, Ideogram-4, and \method{}),
% so every row is scored on an identical set.  Ideogram-4's safety filter refused
% 86 of the 137 prompts (62.8\%); refusals are detected by OCR of the returned
% image containing the substring ``safety filter''.  We report Ideogram-4 both
% over all 137 cases and over the 51 non-blocked cases.

\paragraph{Generation configuration.}
Diffusion baselines and \method{} are run at matched prompts and seeds; the
editor backbone runs in bf16 with classifier-free guidance 5.0 and 40 denoising
steps.  \method{} adds text exactly from the source SVG after
raster synthesis.

\paragraph{Text-fidelity protocol.}
Text fidelity is measured two ways.  \emph{Placement-exact} fidelity uses
ground-truth-region OCR: for each ground-truth text box (taken from the stored
structured conditioning) we crop the corresponding region and OCR it with
Tesseract \cite{tesseractocr2026}.  The primary results use a strict
$768{\times}512$ evaluation canvas. Both exact-text methods are rendered with
\texttt{rsvg-convert \cite{lachowicz2014rsvgconvert}} on the same host with Georgia, Helvetica, and Arial
available through fontconfig.
We report character accuracy ($1-$ normalized edit distance, averaged over
regions) and exact-match rate.  \emph{Placement-agnostic} fidelity instead runs
location-independent full-image OCR and reports word recall, the fraction of
ground-truth content tokens (length $\geq 3$, NFKC-normalized) recovered
anywhere in the image; this credits a baseline that renders legible text at the
wrong position.  For the two exact-text methods the remaining gap from $1.0$
reflects OCR error rather than genuine text mistakes, since the text is placed
by construction.  The two prompt-only baselines behave differently under the two
scores: both are near zero on placement-exact OCR because their glyphs are not
at the ground-truth boxes, but on word recall Ideogram-4 recovers most text
(partially readable yet mislocated) while Qwen-Image-2512 stays low (text is genuinely
garbled).

\paragraph{Judge protocol.}
Aesthetic and typographic quality is scored by a VLM judge using the
human-preference variant of our static-quality rubric.
The primary judge (Claude Opus~4.8)
scores each image on $0$-$5$ across three criteria-\emph{illustration
appeal}, \emph{typography quality}, \emph{semantic coverage} plus a holistic overall score that we omit in the reported tables
in favor of \emph{Avg}, the transparent mean of illustration appeal, typography,
and semantic coverage.  The appeal criterion uses strict, anchored bands so that
coarse, flat, or generic illustrations stay in the low range regardless of how
clean the surrounding layout is, and richly rendered, detailed artwork is
required for the top bands; this keeps the score aligned with what a human would
actually prefer without the rubric explicitly naming photorealism.  The design
concept passed to the judge is reconstructed from the stored structured
conditioning, and judgments are disk-cached for reproducibility.
We additionally repeat the identical image-rubric judgments with Gemini
3.7 Flash \cite{deepmind2026gemini37flash} and GPT-5.6 Sol \cite{openai2026gpt56sol}.  All three judges receive the same $768{\times}512$ resolution.

\paragraph{Full results.}
Table~\ref{tab:eval_full} reports the judge criteria alongside both
text-fidelity metrics.

\begin{table*}[t]
  \caption{Full evaluation, including the \emph{craft} judge
  criterion omitted from Table~\ref{tab:main}. Text fidelity is measured at
  $768{\times}512$ and includes placement-exact
  ground-truth-region OCR (char / exact) and placement-agnostic full-image word
  recall (\emph{Recall}); judge criteria are $0$-$5$, higher
  better; \emph{Avg} is the mean of appeal, typography, and coverage.
  $^\dagger$LLM-authored SVG recolor reference.}
  \label{tab:eval_full}
  \centering
  \small
  \setlength{\tabcolsep}{4.5pt}
  \begin{tabular}{lcccccccc}
    \toprule
    & \multicolumn{3}{c}{Text fidelity} & \multicolumn{5}{c}{VLM judge (0-5)} \\
    \cmidrule(lr){2-4}\cmidrule(lr){5-9}
    Method & Char & Exact & Recall & Appeal & Typog. & Cov. & Craft & Avg \\
    \midrule
    SVG (LLM-authored)$^\dagger$ & 0.856 & 0.676 & 0.964 & 2.37 & 4.02 & 4.90 & 3.91 & 3.76 \\
    \midrule
    Qwen-Image-2512 & 0.080 & 0.002 & 0.149 & 2.69 & 1.00 & 1.45 & 1.51 & 1.71 \\
    Ideogram-4 & 0.657 & 0.360 & 0.744 & 2.92 & 2.29 & 3.89 & 2.49 & 3.03 \\
    \method{} (Ours) & 0.822 & 0.623 & 0.942 & 2.99 & 3.81 & 4.75 & 3.59 & 3.85 \\
    \bottomrule
  \end{tabular}
\end{table*}

\begin{table*}[t]
  \caption{\textbf{OCR resolution sensitivity.} Placement-exact character
  accuracy (\emph{Char}) and exact-match rate (\emph{Exact}), together with
  placement-agnostic word recall (\emph{Recall}).}
  \label{tab:ocr-resolution-sensitivity}
  \centering
  \small
  \setlength{\tabcolsep}{5pt}
  \begin{tabular}{lcccccc}
    \toprule
    & \multicolumn{3}{c}{$768{\times}512$ OCR} &
      \multicolumn{3}{c}{4096-pixel OCR} \\
    \cmidrule(lr){2-4}\cmidrule(lr){5-7}
    Method & Char & Exact & Recall & Char & Exact & Recall \\
    \midrule
    SVG (LLM-authored)$^\dagger$ & 0.856 & 0.676 & 0.964 & 0.883 & 0.595 & 0.814 \\
    \midrule
    Qwen-Image-2512 & 0.080 & 0.002 & 0.149 & 0.077 & 0.002 & 0.150 \\
    Ideogram-4 & 0.657 & 0.360 & 0.744 & 0.643 & 0.268 & 0.727 \\
    \method{} (Ours) & 0.822 & 0.623 & 0.942 & 0.853 & 0.543 & 0.806 \\
    \bottomrule
  \end{tabular}
\end{table*}

\begin{table}[t]
  \caption{\textbf{Multi-judge robustness at $768{\times}512$.} Each entry is
  the mean of illustration appeal, typography quality, and semantic coverage.
  \method{} is the highest-scoring learned synthesis method under every judge.}
  \label{tab:multi-judge-robustness}
  \centering
  \small
  \setlength{\tabcolsep}{4pt}
  \begin{tabular}{lcccc}
    \toprule
    Judge & SVG & Qwen & Ideogram & \method{} \\
    \midrule
    Claude Opus 4.8 \cite{anthropic2026claudeopus48} & 3.76 & 1.71 & 3.03 & \textbf{3.85} \\
    Gemini 3.7 Flash \cite{deepmind2026gemini37flash} & 3.93 & 1.99 & 3.54 & \textbf{3.99} \\
    GPT-5.6 Sol \cite{openai2026gpt56sol} & 3.81 & 1.99 & 3.32 & \textbf{3.81} \\
    \bottomrule
  \end{tabular}
\end{table}

Across judges, \method{}'s Avg is $3.88\pm0.08$ (population standard
deviation), compared with $1.90\pm0.13$ for Qwen and $3.30\pm0.21$ for
Ideogram.  This slightly exceeds the SVG reference's
$3.83\pm0.07$ cross-judge mean.  Judges differ in absolute calibration but
preserve the learned-method ranking.

\subsection{Additional comparisons}
\label{app:additional-comparisons}

\begin{figure*}[p]
  \centering
  \includegraphics[width=\linewidth]{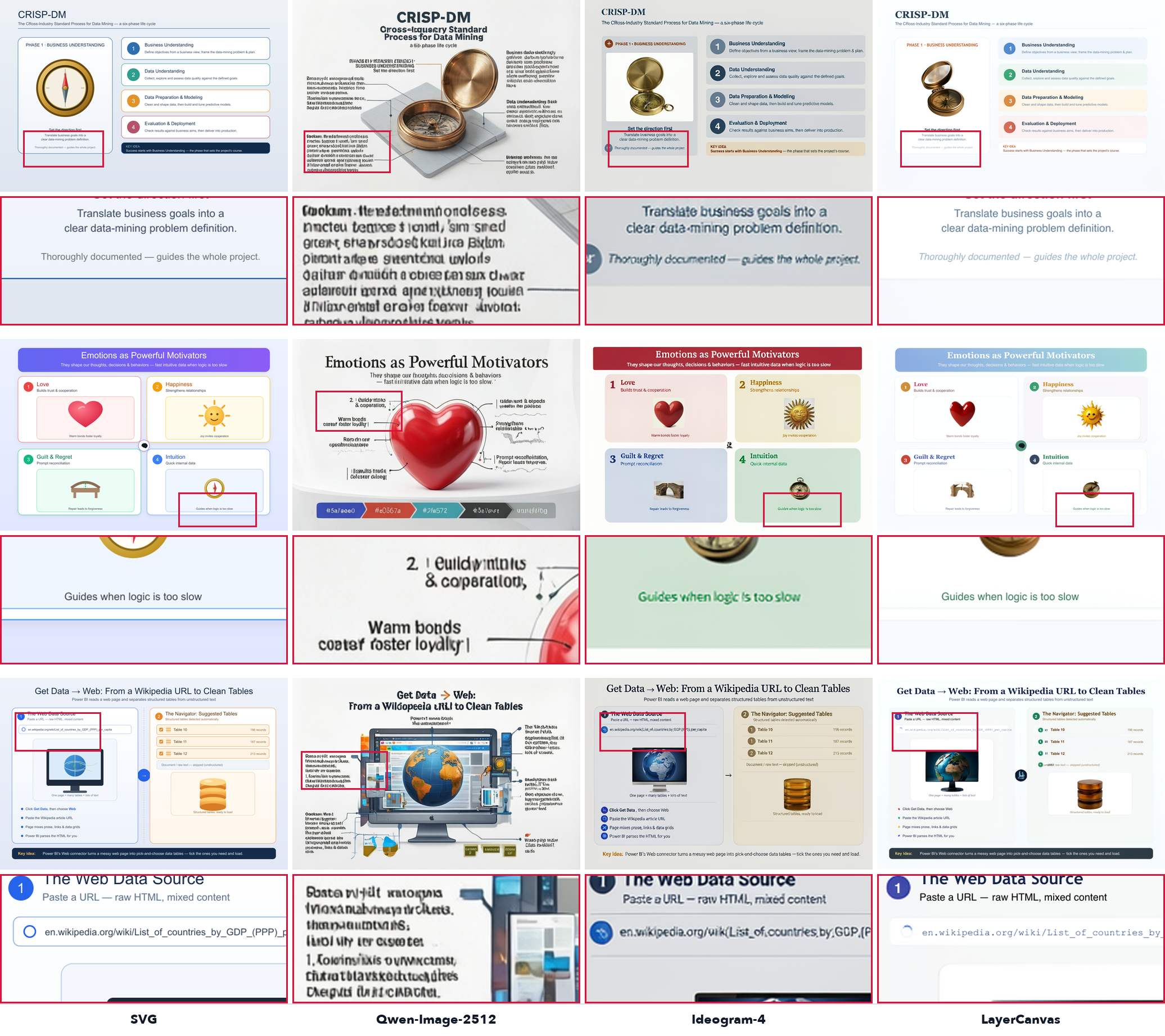}
  \caption{\textbf{Additional qualitative comparisons.}
  Columns show the LLM-authored SVG reference, Qwen-Image-2512, Ideogram-4,
  and \method{}, respectively.}
  \label{fig:additional-qualitative-comparisons-1}
\end{figure*}

\begin{figure*}[p]
  \centering
  \includegraphics[width=\linewidth]{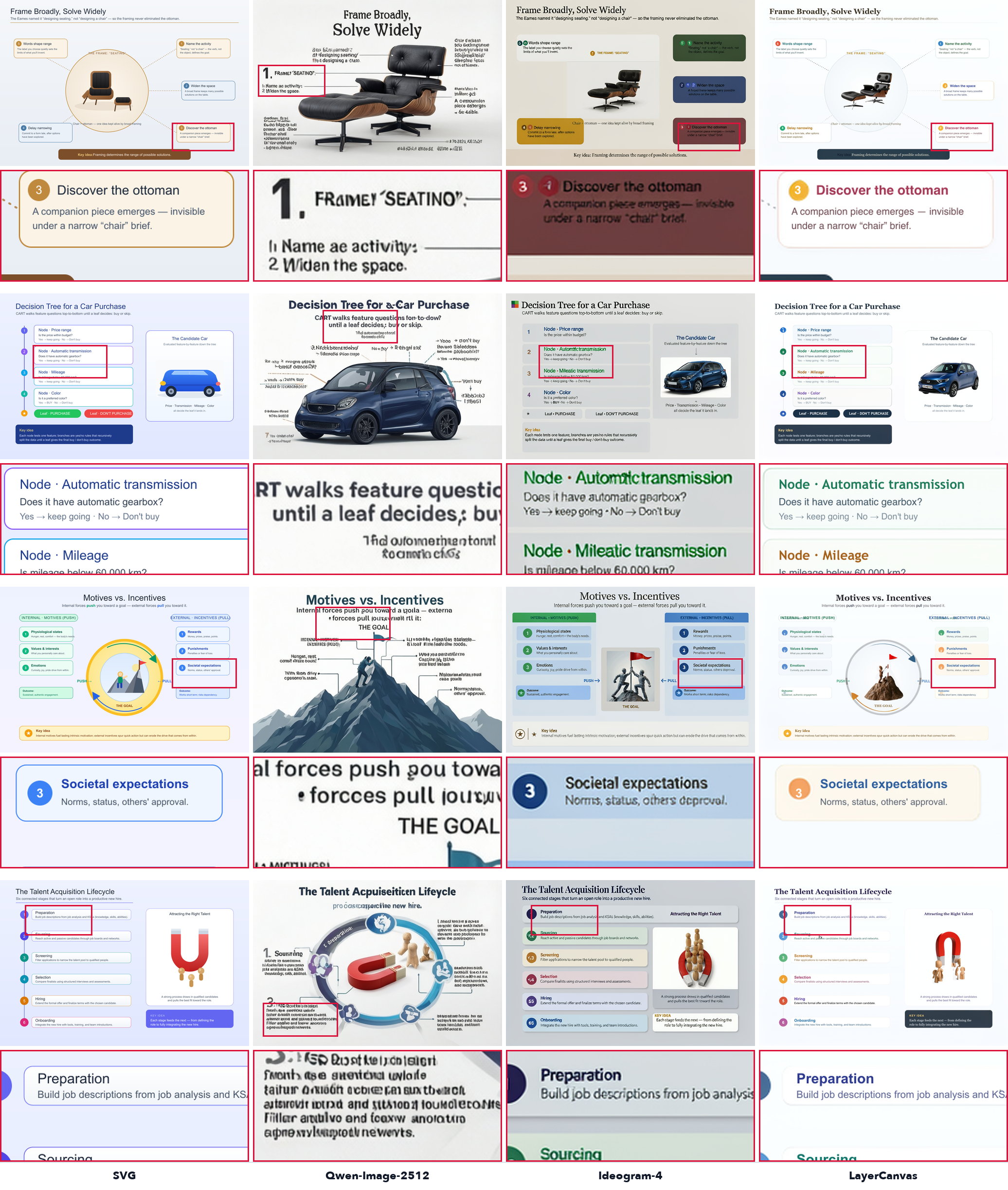}
  \caption{\textbf{Additional qualitative comparisons.}
  Columns show the LLM-authored SVG reference, Qwen-Image-2512, Ideogram-4,
  and \method{}, respectively.}
  \label{fig:additional-qualitative-comparisons-2}
\end{figure*}

Figures~\ref{fig:additional-qualitative-comparisons-1} and~\ref{fig:additional-qualitative-comparisons-2} provide additional qualitative comparisons, showing that \method{} preserves the source layout and legible text while enriching the illustrations and avoiding the text failures of raster baselines.

\subsection{Systems profile}
\label{app:systems-profile}

\subsubsection{Training compute}
\label{app:training-compute}

With full gradient checkpointing, a single-H200 training step takes
10.33\,s and uses 30.8\,GiB of peak GPU memory. Backward recomputation takes
7.18\,s, forward execution 2.29\,s, and input preparation 0.83\,s.
Checkpointing half of the transformer blocks reduces the step time to
approximately 9.2\,s but raises peak GPU memory to approximately 108\,GiB.
Quarter and no-checkpointing profiles exceed 133\,GiB and run out of memory on
a 141\,GB H200.

\subsubsection{Inference compute}
\label{app:inference-compute}

\begin{figure}[t]
  \centering
  \includegraphics[width=0.92\linewidth]{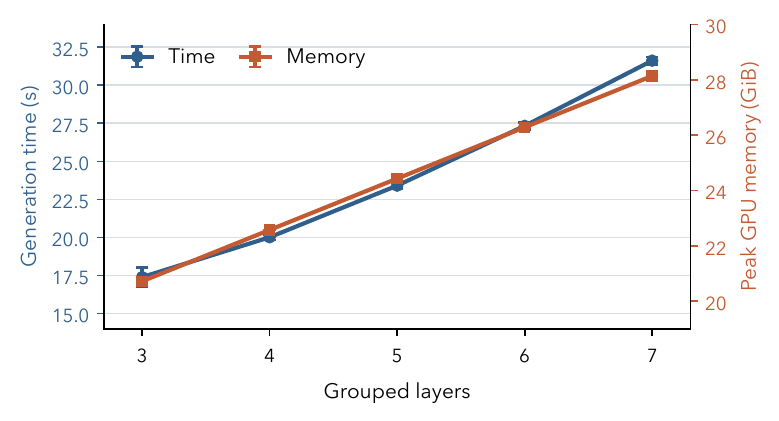}
  \caption{\textbf{Diffusion inference compute.} Generation time and peak
  allocated GPU memory on one H200 increase with the number of grouped layers.
  Each point is the mean of two 20-step generations; error bars show the
  measured minimum and maximum.}
  \label{fig:inference-time-vs-layers}
\end{figure}

\paragraph{Layer generation.}
A matched single-H200 profile uses two validation examples at each grouped
layer count from three through seven. End-to-end generation takes 24.0\,s per
sample on average and 23.4\,s at the median, with a measured range of
16.8-31.9\,s over 10 generations. Peak allocated GPU memory rises from
20.7\,GiB at three grouped layers to 28.1\,GiB at seven. The trend is shown in Figure \ref{fig:inference-time-vs-layers}.

\clearpage
\section{Downstream Applications}
\label{app:downstream-applications}

This section demonstrates how the hybrid raster-SVG layer-set representation
supports deterministic editing and reuse after generation. Each application uses
the trained grouped layer-set model or a previously generated output from that
model; operations performed only on the persistent structured representation are
identified explicitly.

\subsection{Text editing and localization}
\label{app:downstream-text-editing-localization}

End-to-end raster generators must repaint text whenever wording, typography, or
language changes. Because their glyph generation is data-driven, they may also
favor the visual conventions of a dominant training language rather than
faithfully rendering a requested, less represented script. \method{} instead
generates only the visual substrate and preserves typography as styled SVG
nodes. Copy, style, and locale changes therefore require no diffusion pass and
cannot alter the generated artwork.

\begin{figure}[t]
  \centering
  \includegraphics[width=\linewidth]{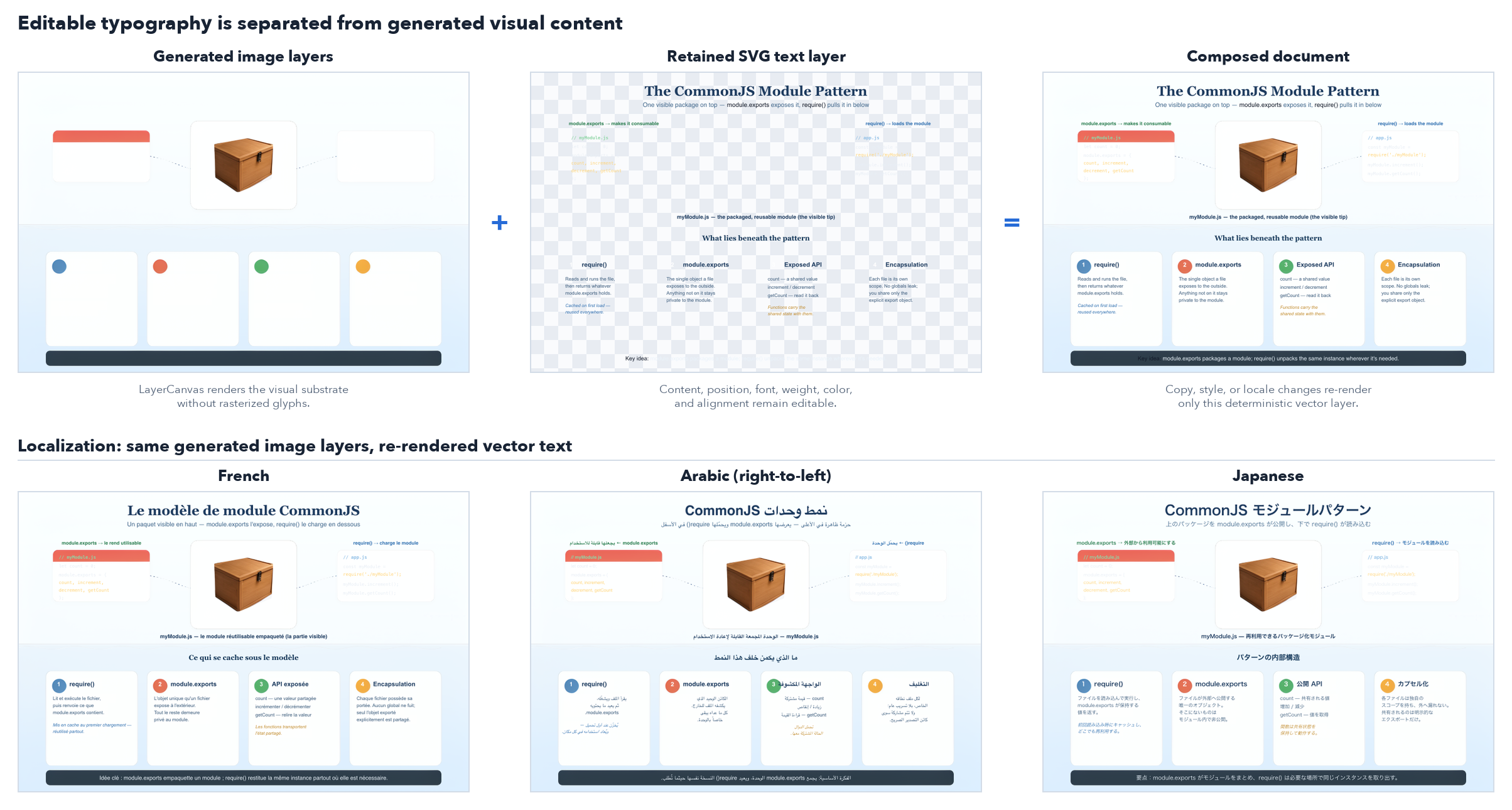}
  \caption{\textbf{Editable text representation and localization.} A held-out
  LayerCanvas result is generated without glyphs, while the source SVG text
  nodes remain a separate editable vector layer. Their composition produces the
  final document. The lower row localizes the same unchanged generated image
  layers into French, Arabic, and Japanese by re-rendering only the text layer.}
  \label{fig:downstream-text-editing-localization}
\end{figure}

\paragraph{Implementation.}
Each text node retains its Unicode content, position, font family, size, weight,
style, anchor, fill color, transforms, and parent ownership. Editing replaces
only the requested attributes before the text-only SVG is rasterized and
alpha-composited over the unchanged model output. Localization uses a
script-appropriate local font and shaping direction; nonlinguistic values such
as numbers and code may remain unchanged. Figure~\ref{fig:downstream-text-editing-localization}
shows the same generated layers with text rendered in four languages.

\paragraph{Deterministic behavior.}
This operation is a property of the representation rather than a learned
prediction: the requested node attributes are written directly into SVG and
rendered by the vector-text runtime. Thus the rendered language follows the
provided Unicode content and selected font rather than a language distribution
learned by the image model. The only external requirements are an available font
with the requested glyphs and enough space in the retained text box for the new
copy. If a translation changes line length substantially, standard SVG layout
operations can resize or reflow the text box without regenerating any image
layer.

\subsection{Independent layer animation}
\label{app:downstream-animation-progressive-reveal}

We adapt the zero-shot animation principle of
Text2Video-Zero~\cite{khachatryan2023text2videozero}, which shows that the motion
and appearance priors of a pretrained text-to-image diffusion model can produce
a temporally linked sequence without video-specific training. A flattened
document would still expose the complete page to this generative update, allowing
unrelated illustrations, layout, and typography to drift. LayerCanvas instead
lets the temporal operation target one named image layer while all other layers
remain fixed.

\paragraph{Implementation.}
Following Text2Video-Zero, we derive a short motion trajectory from an image
diffusion prior rather than training a video model. The user selects an image
layer, optionally together with a bounded subregion, and specifies endpoint
states or a motion transform. Intermediate states are produced through latent
interpolation or motion-guided warping, decoded only for the selected content,
and alpha-composited over the unchanged layer set at each timestep. This final
step is specific to our representation: every unselected image layer and the
full-resolution SVG typography are copied identically into every frame. In
Figure~\ref{fig:downstream-animation-progressive-reveal}, the selected subregion
is an eye band covering $6.6\%$ of the canvas.

\paragraph{Discussion.}
The generative component supplies motion only; spatial isolation comes from the
LayerCanvas representation. Pixels outside the selected region and the complete
vector-text layer are not predicted at subsequent timesteps, so they remain
bit-identical by construction. This converts a global zero-shot image animation
method into a local document operation and avoids spending temporal model
capacity on content that should not move.

\subsection{Asset reuse and recomposition}
\label{app:downstream-asset-reuse-recomposition}

A flattened image binds every illustration, diagram, and decorative element to
its original page. Reusing one of them requires segmentation followed by
inpainting or regeneration of the surrounding content. LayerCanvas instead keeps
generated visual groups as addressable image layers with their own alpha and
paint order, so an element can become a reusable asset after generation.

\begin{figure}[t]
  \centering
  \includegraphics[width=\linewidth]{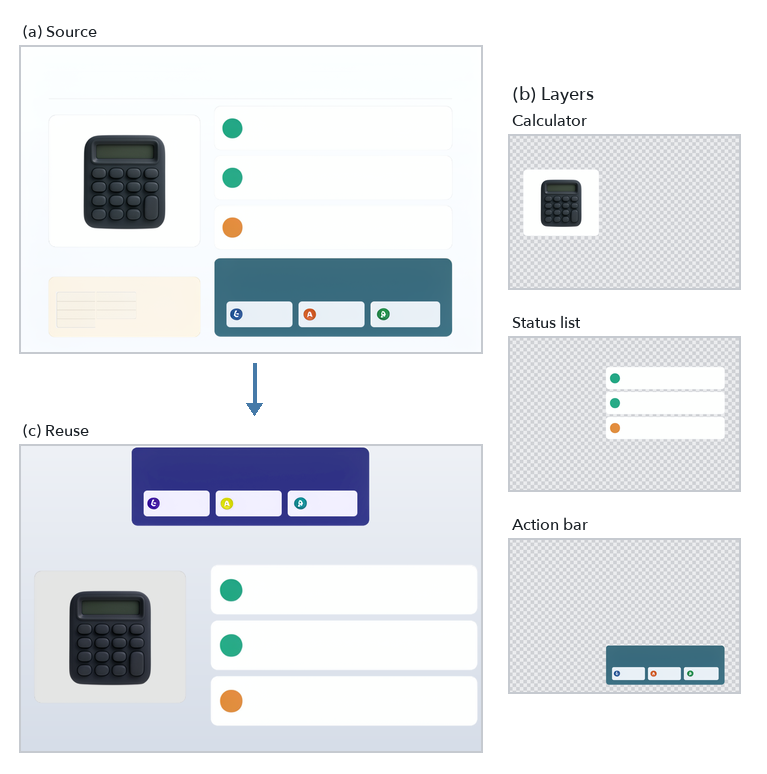}
  \caption{\textbf{Asset reuse.} Three generated layers are taken from one
  document and arranged into a new composition. Each layer carries its own
  alpha, allowing independent repositioning, scaling, restyling, and reordering.}
  \label{fig:downstream-asset-reuse-recomposition}
\end{figure}

\paragraph{Implementation.}
A selected layer is copied with its image content and alpha mask. Standard
layer operations then translate or scale it, adjust its appearance within the
mask, and assign a new position in the target paint order. Multiple selected
layers may be placed onto a new background and combined with newly authored SVG
text. These operations act on the saved representation and do not require
segmentation, inpainting, or another diffusion pass. Figure~\ref{fig:downstream-asset-reuse-recomposition}
shows three layers reused in a new composition.

\paragraph{Discussion.}
Reuse preserves the generated appearance of an asset while decoupling it from
the document in which it first appeared. A collection of generated layers can
therefore serve as a project-specific asset library: recurring illustrations or
diagram components may be placed across related documents while layout and text
remain independently editable. The resulting composition is still structured,
so the reused layer can be moved, hidden, or replaced again later.

\subsection{Layout retargeting and responsive resize}
\label{app:downstream-layout-retargeting}

A design may need to appear as a presentation slide, poster, and mobile graphic.
Resizing a flattened page either distorts its elements or removes content through
cropping. LayerCanvas retains the document as addressable groups, allowing the
same generated content to be arranged for a different canvas.

\begin{figure}[t]
  \centering
  \includegraphics[width=\linewidth]{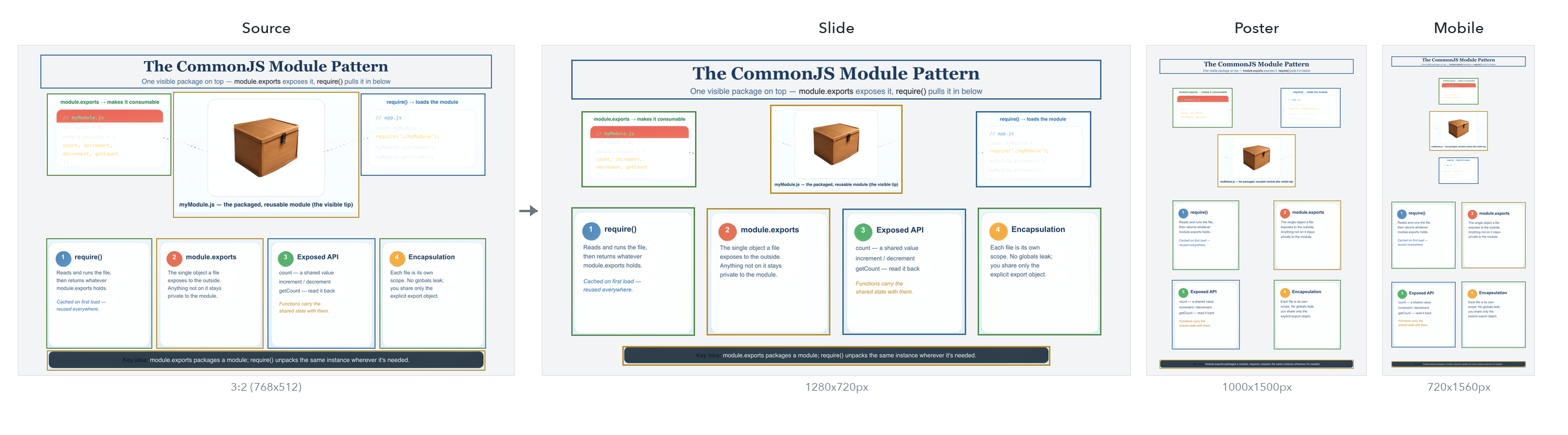}
  \caption{\textbf{Layout retargeting.} The same generated module illustration,
  code panels, and explanatory cards are rearranged for slide, poster, and
  mobile canvases. Image groups preserve their aspect ratios, while text remains
  separately renderable for each target layout.}
  \label{fig:downstream-layout-retargeting}
\end{figure}

\paragraph{Implementation.}
Each illustration, diagram, card, and text block carries a bounding box and
group assignment. A simple reflow solver packs these groups in reading order,
using more columns on wide canvases and a vertical stack on narrow ones.
Generated image groups are scaled uniformly, while SVG text can be re-rendered
for the new box. The operation changes only layout metadata and recomposition;
the image layers are not generated again. Figure~\ref{fig:downstream-layout-retargeting}
shows the same content adapted to slide, poster, and mobile layouts.

\paragraph{Discussion.}
Retargeting becomes a document-layout problem rather than an image-generation
problem. Groups may be reprioritized, resized, or moved between rows without
changing their generated appearance, and the result remains editable after
recomposition. More advanced layout objectives or design templates can replace
the simple solver without changing the underlying representation.

\subsection{Command-driven interactive editing}
\label{app:downstream-command-driven-editing}

Interactive authoring is naturally expressed through commands such as ``remove
this card,'' ``emphasize this panel,'' or ``swap this icon.'' Prompt-based
regeneration may redraw the entire canvas. Named layer groups instead allow the
command to be grounded to one element and executed as a local, reversible edit.

\begin{figure}[t]
  \centering
  \includegraphics[width=\linewidth]{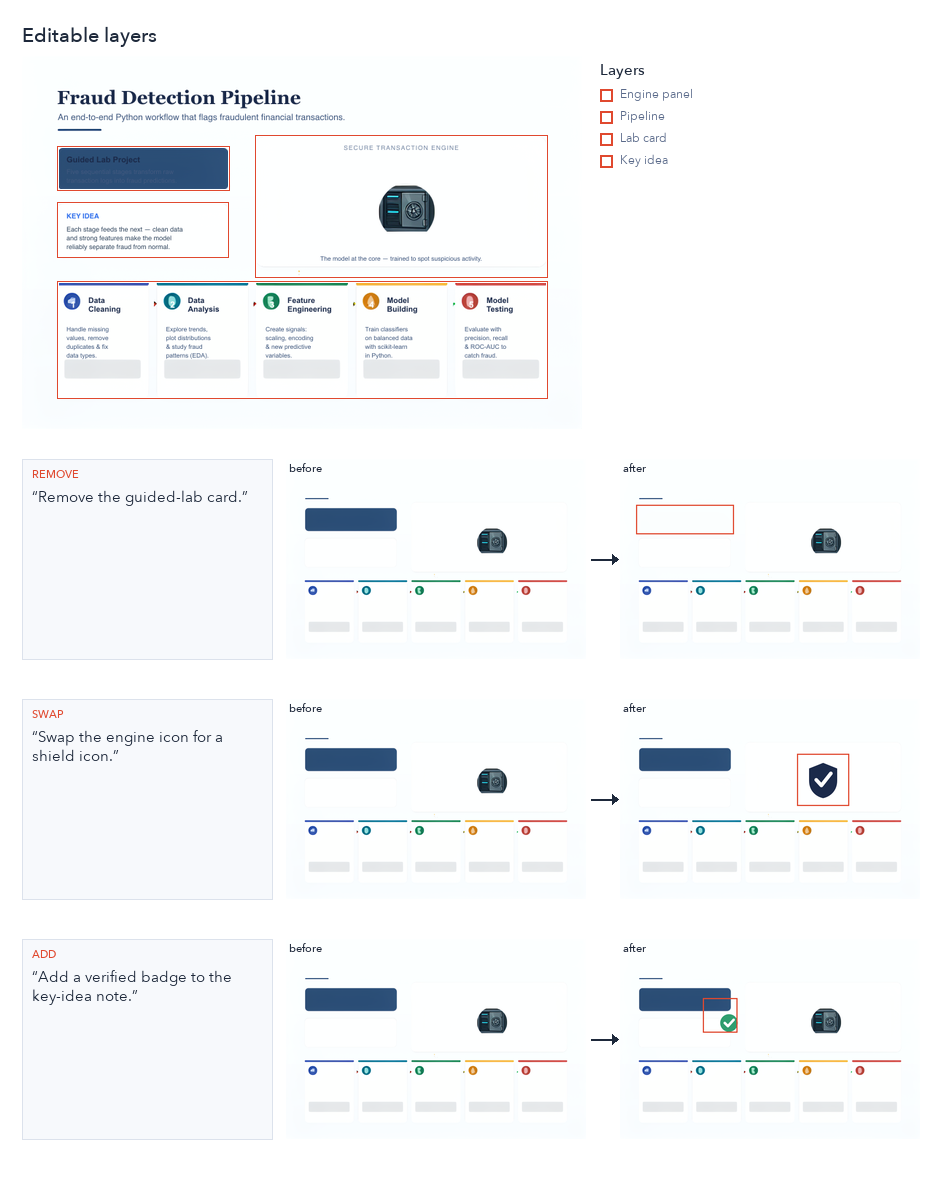}
  \caption{\textbf{Command-driven editing.} Three natural-language commands are
  converted to targeted remove, swap, and add operations. Red boxes indicate
  the edited regions; unselected layers are reused unchanged.}
  \label{fig:downstream-command-driven-editing}
\end{figure}

\paragraph{Implementation.}
A lightweight interpreter maps a command to
$\{\textsc{action},\textsc{target}\}$ using action phrases and aliases for the
document's semantic groups. We support removing, emphasizing, swapping, and
adding elements. The executor applies the operation to the targeted SVG group or
model-rendered layer. Full-resolution alpha masks define the affected region,
allowing the result to be recomposited without resampling unrelated content.
Figure~\ref{fig:downstream-command-driven-editing} shows remove, swap, and add
commands applied to selected regions.

\paragraph{Discussion.}
The language interface does not edit pixels directly. It selects an operation
and resolves a semantic target, after which the persistent layer representation
defines the exact region and compositing behavior. Commands can therefore be
previewed, reversed, or chained without resampling the full document. A learned
language parser could replace the lightweight interpreter without changing the
underlying editing operations or their locality.

\subsection{SVG-space refinement and faithful diffusion propagation}
\label{app:downstream-svg-refinement-propagation}

An upstream layout proposal may contain a misplaced, mislabeled, or incorrectly
styled element. In a raster-only pipeline this often requires a new prompt or
manual pixel editing. In \method{}, the SVG is a persistent control surface:
the element can be corrected directly, and the model propagates that correction
to the raster without retraining or redesigning the rest of the document.

\paragraph{Implementation.}
We deliberately perturb one SVG element by translating, resizing, recoloring,
relabeling, or replacing it. The incorrect and corrected conditions use the same
sampling noise so that differences are attributable to the structural edit.
Composition uses each group's full-resolution vector alpha mask rather than an
upsampled latent-grid mask. Figure~\ref{fig:downstream-svg-refinement-propagation}
presents the incorrect condition, the SVG edit, and the corrected render as a
matched triptych.

\paragraph{Discussion.}
The structured intermediate remains editable after an initial generation rather
than becoming a disposable prompt. A user or upstream layout system can inspect
the result, correct the responsible SVG group, and regenerate the affected
visual content under the revised condition. Fixed noise helps preserve the
remaining appearance, while recomputing masks and overlap relations allows
changes in position, scale, or occlusion to propagate consistently.

\clearpage
\section{Limitations, ethics, and broader impact.}
\label{app:limitations}
Despite these benefits, LayerCanvas requires a valid source SVG and inherits errors and biases in its content, layout, and training data. Deterministic typography preserves supplied text but does not ensure factuality, accessibility, or culturally appropriate localization, while edits that change occlusion or z-order may affect unintended regions. Responsible deployment therefore requires documented data provenance, license and bias checks, generated-content disclosure, and human review.

\end{document}